\documentclass{article} 
\usepackage{iclr2027_conference_custom,times}

\usepackage{algorithm}
\usepackage{algpseudocode}
\usepackage{graphicx}
\usepackage[table]{xcolor}
\usepackage{amsmath,amssymb}
\usepackage{booktabs}
\usepackage{hyperref}
\usepackage{url}
\usepackage{wrapfig}
\usepackage{caption}
\usepackage{enumitem}

\usepackage{amsmath,amsfonts,bm}

\def\eqref#1{equation~\ref{#1}}

\def\1{\bm{1}}

\def\vzero{{\bm{0}}}
\def\vone{{\bm{1}}}

\def\ve{{\bm{e}}}
\def\vf{{\bm{f}}}

\def\vq{{\bm{q}}}
\def\vr{{\bm{r}}}

\def\vu{{\bm{u}}}
\def\vv{{\bm{v}}}

\def\vx{{\bm{x}}}

\def\mI{{\bm{I}}}
\def\mJ{{\bm{J}}}

\def\mM{{\bm{M}}}

\def\mQ{{\bm{Q}}}
\def\mR{{\bm{R}}}

\def\mV{{\bm{V}}}
\def\mW{{\bm{W}}}

\DeclareMathAlphabet{\mathsfit}{\encodingdefault}{\sfdefault}{m}{sl}
\SetMathAlphabet{\mathsfit}{bold}{\encodingdefault}{\sfdefault}{bx}{n}

\newcommand{\vepsilon}{\boldsymbol{\epsilon}}
\newcommand{\normal}{\mathcal{N}}

\algnewcommand\LineComment[1]{\Statex {\color[HTML]{228B22} \hspace{-1.5em} \# #1}}

\newcommand{\blfootnote}[1]{%
  \begingroup
  \renewcommand{\thefootnote}{}
  \def\thefootnotemark{}
  \makeatletter
  \makeatother
  \footnotetext{#1}
  \endgroup
}

\title{On the spectral properties of generative denoiser Jacobians}

\author{Alexandros Graikos \\
Stony Brook University\\
\And
Nebojsa Jojic \\
Microsoft Research \\
\And
Dimitris Samaras \\
Stony Brook University\\
}

\iclrfinalcopy

\begin{document}

\maketitle

\begin{abstract}
Generative denoising models, such as diffusion and flow-matching, learn to sample from complex distributions by training a deep neural network denoiser to recover clean data from noise-corrupted samples. While such models are typically compared on the quality of their synthesized samples, these metrics provide limited insight into how the underlying denoiser, which drives generation, differs. In this work, we propose to analyze the spectrum of the denoiser Jacobian as a tool to characterize these differences. Across pre-trained denoising models, we observe that better generative performance is associated with larger Jacobian eigenvalues. Motivated by this, we introduce a regularization scheme that controls the Jacobian spectrum by training the denoiser on perturbed inputs, with perturbations suppressing or amplifying Jacobian responses. On ImageNet, we test whether directly modifying the Jacobian spectral properties leads to improved generations. Our findings suggest that denoisers benefit from both strengthening responses along data-relevant principal eigen-directions and suppressing the noisy, data-irrelevant ones. This establishes the denoiser Jacobian as a useful tool for identifying differences between generative denoising models.
\blfootnote{Corresponding author agraikos@cs.stonybrook.edu.}
\blfootnote{Code provided at \href{https://github.com/AlexGraikos/denoiser_jacobian_spectrum}{this repository}.}
\end{abstract}

\section{Introduction}
\label{sec:intro}

Denoising generative models, such as diffusion \citep{ho2020denoising} and flow-matching \citep{lipman2023flow}, learn to sample from complex distributions using a deep neural network denoiser. For natural images, the denoiser looks at noise-corrupted samples and utilizes patterns learned from the training data to predict the underlying clean image. By repeatedly making such predictions, denoising models gradually transform random Gaussian noise into realistic images \citep{song2021denoising}.

Once trained, the typical approach for comparing models is to evaluate the quality of their generations. While the literature has developed a multitude of metrics \citep{heusel2017gans,sajjadi2018assessing}, these ultimately only characterize differences in model outputs, therefore providing limited insight into how the learned denoising function differs, e.g., the patterns used in denoising images. The question then is \emph{why} one denoiser synthesizes samples better than the other, and since sample quality alone cannot answer it, we turn to the properties of the denoising function itself.

In this work, we examine the Jacobian of the learned denoiser as a means to probe its internal properties and reveal differences across models. The Jacobian specifies how small changes in the noisy input map to changes in the denoised output, which inherently captures the dependencies learned by the trained model \citep{graikos2026fast,lukoianov2026locality}. For instance, when perturbing a single noisy input pixel, the Jacobian reveals which pixels in the output are affected and in what way, encoding the covariances used to effectively denoise samples.

Since the Jacobian acts as a learned covariance matrix, we turn to spectral analysis, i.e., eigenvectors and eigenvalues, to describe its behavior. Starting from pre-trained generative models (SiTs \citep{ma2024sit}), we uncover an intriguing relationship: models that are better in generative metrics also exhibit larger denoiser Jacobian eigenvalues, which correspond to sharper responses of the learned denoising function along the eigen-directions.

Motivated by this observation, we next ask whether the reverse also holds: does altering the Jacobian spectrum change generation quality? We first introduce a simple regularization scheme that controls the Jacobian spectrum by perturbing the denoiser input during training. We show that, with an appropriate choice of perturbation, we can either suppress spurious directions in the Jacobian spectrum or encourage larger gain along principal eigenvectors, without ever explicitly computing the eigendecomposition during training.

Finally, we apply this regularization to ImageNet models, a setting comparable to the pre-trained model analysis. Our findings show that spectral regularization improves generative metrics both when encouraging large responses along eigen-directions and when suppressing noisy, data-irrelevant responses. We further show that these two objectives are composable, yielding cumulative improvements when applied together and suggesting that an effective model should exhibit both properties in its Jacobian spectrum.

By providing a framework that pinpoints differences between models through their Jacobian spectra, we target a more complete view of how training shapes the learned denoising function. With regularization becoming increasingly popular for training speed and generation quality \citep{yu2025representation}, but yet partially understood \citep{singh2025matters}, any principled effort to describe these differences is crucial, both for understanding existing methods and designing future objectives. Our results in this work suggest that the denoiser Jacobian is a useful tool for doing so.
We summarize our contributions as follows:
\begin{itemize}[leftmargin=*]
    \item We establish spectral analysis as a useful tool for revealing differences between generative denoising models. Specifically, we show that better-performing models exhibit larger eigenvalues in their denoiser Jacobian, uncovering a previously unknown relationship between Jacobian spectrum and generative performance.
    \item We propose a regularization scheme that controls the denoiser Jacobian by perturbing model inputs during training. For different perturbations, we show how we can selectively suppress or amplify the Jacobian response along random or eigen-directions.
    \item We apply the proposed regularizer on ImageNet-trained models and demonstrate that generative performance improves when increasing principal Jacobian eigenvalues. We further show that this effect is complementary to suppressing the noisy, data-irrelevant responses, with the two yielding cumulative improvements when combined.
\end{itemize}

\section{Related Work}
\label{sec:related_work}

\textbf{Denoising generative models}
Denoising is the primary approach for learning to generate physical signals (e.g., images \cite{rombach2022high,esser2024scaling}, audio \citep{evans2024fast}). Although the idea of modeling data distributions by denoising is not new \citep{bengio2013generalized,sohl2015deep}, recent advances have produced capable generative models with both algorithmic \citep{ho2020denoising,song2021scorebased,lipman2023flow} and architectural improvements \citep{nichol2021improved,peebles2023scalable,ma2024sit}. Regardless of specific formulation, every model is built using a learned denoiser, whose spectral properties are the focus of this work.

\textbf{Denoiser Jacobians and spectral analysis.}
Recent work has looked at the Jacobian of the learned denoiser, which reveals dependencies between input and output pixels, to probe the learned statistics of the data distribution \citep{lukoianov2026locality,graikos2026fast}. Regarding spectral properties, \citet{manor2024posterior} related the posterior uncertainty to the Jacobian in (non-generative) denoisers and employed a similar power iteration algorithm to compute principal components. \citet{kadkhodaie2024generalization} showed that eigenvectors of a generative denoiser form a data-adaptive harmonic basis. While these works motivate spectral analyses of the Jacobian, they mostly focus on individual models. In contrast, we draw comparisons across different denoisers, which only then reveals how spectrum relates to generation quality.

\textbf{Training regularization}
Regularizing the training of diffusion and flow-matching models has recently garnered widespread attention, particularly with the success of Representation Alignment (REPA \citep{yu2025representation}). By showing that it accelerates training and benefits generation quality, regularization offers an appealing research direction for improving generative models. This has motivated works that regularize with a student-teacher framework \citep{chefer2026selfsupervised}, or by utilizing contrastive objectives \citep{stoica2025contrastive,wang2025diffuse}.

Closer to our work, \citet{ning2023input} perturb denoiser inputs during training with Gaussian noise to mitigate the mismatch between inputs seen in training and sampling, while \citet{scarvelis2024nuclear} showed that perturbations can impose nuclear norm regularization on the Jacobian without explicitly constructing it. Interestingly, \citet{alain2014regularized} previously connected denoising and Jacobian regularization to the local geometry of the data distribution, which coincides with the effect of random perturbations. Compared to previous works, which suppress the Jacobian, we are the first to explicitly regularize for an \emph{increased} Jacobian response.

\section{Spectral properties of denoiser Jacobians}
\label{sec:spectral_properties}

\subsection{Generative denoising models}
\label{sec:generative_denoising_models}

We use the term \emph{generative denoising models} to refer to both diffusion \citep{ho2020denoising} and flow-matching \citep{lipman2023flow}. These models learn to transform Gaussian noise $\vepsilon \sim \normal(\vzero, \mI)$ into data $\vx_0 \sim p(\vx_0)$ using a trained denoiser network that recovers information from noisy samples $\vx_t$. In this work, we consider denoisers operating in the linear noise schedule \citep{ma2024sit}
\begin{equation}
    \vx_t = t\vx_0 + (1-t)\vepsilon,\quad t \in [0,1].
    \label{eq:forward_q}
\end{equation}
To synthesize new data we train the denoiser $\vf_{\theta}$ using the objective
\begin{equation}
    \mathcal{L}(\theta) = w(t) \left\lVert \vf_{\theta}(\vx_t,t) - \vx_0 \right\rVert_2^2
    \label{eq:denoising}
\end{equation}
for randomly sampled batches $(\vx_0, \vx_t, t)$ of clean and noisy samples.
We draw new samples, starting from $\vx_1 \sim \normal(\vzero, \mI)$, and numerically solving the ODE
\begin{equation}
    \frac{d\vx_t}{dt} = \vu_{\theta}(\vx_t,t),\quad
    \vu_{\theta}(\vx_t,t)
    = \frac{\vf_{\theta}(\vx_t,t)-\vx_t}{1-t}
    \label{eq:ode_sampling}
\end{equation}
with the solver operating backwards in time to produce $\vx_0$  from the implicit distribution $p_{\theta}(\vx_0)$ learned by the trained denoiser $\vf_{\theta}$. 

Some design choices vary across different denoising models, such as noise schedule \citep{karras2022elucidating,chen2023importance,esser2024scaling}, velocity-prediction $\vu$ instead of $\vx_0$ \citep{li2026back}, and the per-timestep weighting $w(t)$ \citep{karras2022elucidating}. Nevertheless, we can always obtain the denoising function $\vf_{\theta}(\vx_t,t)$, which is our object of interest.

\subsection{Denoiser Jacobian properties}
\label{sec:spectral_analysis}

The denoiser Jacobian matrix $\mJ_{\theta}(\vx_t,t) = {\partial \vf_{\theta}(\vx_t,t)}/{\partial \vx_t}$ maps local changes in the noisy input $\vx_t$ to changes in the ``clean'' output $\vf_{\theta}(\vx_t,t)$. These capture the learned \emph{covariances} at each noise level; for instance, when changing a single pixel in the noisy input image, the Jacobian highlights the affected output pixels, revealing correlations learned by the model for the given image. This property is formally expressed by the relationship between the Jacobian and posterior covariance
\begin{equation}
    \mJ = \frac{t}{(1-t)^2} \operatorname{Cov}(\vx_0 \mid \vx_t).
    \label{eq:jac_covariance}
\end{equation}
For a derivation, refer to Appendix~\ref{sec:jac_covariance}. Although the identity only applies to the ideal denoiser case \citep{efron2011tweedie}, it prompts us to analyze the Jacobian with tools used for covariance matrices, i.e. \emph{spectral analysis}. By looking at the properties of the Jacobian, we examine the first-order characteristics of the denoiser (differences in denoising), which can potentially provide more information than a zeroth-order analysis (denoising).

The spectral analysis of the Jacobian involves computing the principal components, or in other words, the eigenvectors corresponding to the directions of largest variance and the eigenvalues measuring expansion along each eigen-direction. Although the Jacobian of a learned denoiser is not perfectly symmetric or positive semi-definite, we will refer to the principal components as eigenvectors and consider the eigenvalues to be real and positive. For the image denoisers we study, the eigenvectors themselves are images that show the maximal changes we can get in the output image by perturbing the noisy input.

The classical numerical method to compute eigenvectors and eigenvalues is power iteration, where we iteratively multiply a probe vector with the target matrix and re-normalize as $\vv_{i+1} \gets {\mJ_{\theta}\vv_i}/{\lVert \mJ_{\theta}\vv_i \rVert}$. Instead of a single probe that only recovers the largest eigenvector, we can use a simultaneous (or block) iteration \citep{clint1970evaluation} to extract the top-$n$ directions.

In the case of the deep neural network denoisers we consider, computing Jacobian-vector multiplications $\mJ_{\theta}\vv_i$ is non-trivial as dimensions increase. To speed up the Jacobian-vector product computations, we use the finite-difference approximation along a direction $\vv$ 
\begin{equation}
    \mJ_{\theta}\vv \approx \frac{\vf_{\theta}(\vx_t+\delta \vv,t)-\vf_{\theta}(\vx_t,t)}{\delta}.
    \label{eq:finite_difference}
\end{equation}
This approximation is shown to be effective and provides a fast and accurate alternative to auto-differentiation \citep{graikos2026fast}. Putting together the simultaneous power iteration and the finite-difference approximation, we propose Algorithm~\ref{alg:simultaneous_iteration}. We use this algorithm as the tool to find the top-$n$ eigenvectors and corresponding eigenvalues of any denoiser Jacobian matrix $\mJ_{\theta}$. Importantly, this algorithm only requires forward passes through the model that can be parallelized across eigenvectors, making it tractable to run even for large $n$ (Appendix~\ref{sec:additional_results}).

\begin{algorithm}[t]
\caption{Simultaneous iteration to compute top-$n$ eigenvectors and eigenvalues of $\mJ_{\theta}$.}
    \textbf{Input}: Noisy sample $\vx_t$, denoiser $\vf_{\theta}(\vx_t,t)$, 
    \verb|#| of eigenvectors $n$, \verb|#| of iterations $K$.
    \begin{algorithmic}[1]
    \State Starting basis
    $\mV^{(0)} = [ \vv_1^{(0)}\ \dots\ \vv_n^{(0)} ]$
    in $\mathbb{R}^n$,
    $\vv_i^{(0)} \sim \normal(\mathbf{0}, \mI)$
    \State Obtain the factors
    $\mQ^{(0)}\mR^{(0)} = \mV^{(0)}$
    \Comment{QR decomposition}
    \For{$k=1,2,\dotsc, K$}
        \State $\mW \gets \mJ \mQ^{(k-1)} 
        = [ \mJ\vq_1^{(k-1)} \dots\ \mJ\vq_n^{(k-1)} ]$
        \Comment{$\mJ\vq_i \approx
        (\vf_{\theta}(\vx_t+\delta \vq_i,t)-\vf_{\theta}(\vx_t,t))/\delta$}
        \State Obtain the factors
        $\mQ^{(k)}\mR^{(k)} = \mW$
    \EndFor
    \end{algorithmic}
    \textbf{Return} eigenvectors $ \mQ^{(K)} = \mV^{(K)} = [ \vv_1^{(K)} \dots\ \vv_n^{(K)} ]$,
    eigenvalues $\lVert \mJ\vv_1^{(K)} \rVert_2,\ \dotsc,\ \lVert \mJ\vv_n^{(K)} \rVert_2$
\label{alg:simultaneous_iteration}
\end{algorithm}

\begin{figure}[t]
    \centering
    \includegraphics[width=\linewidth]{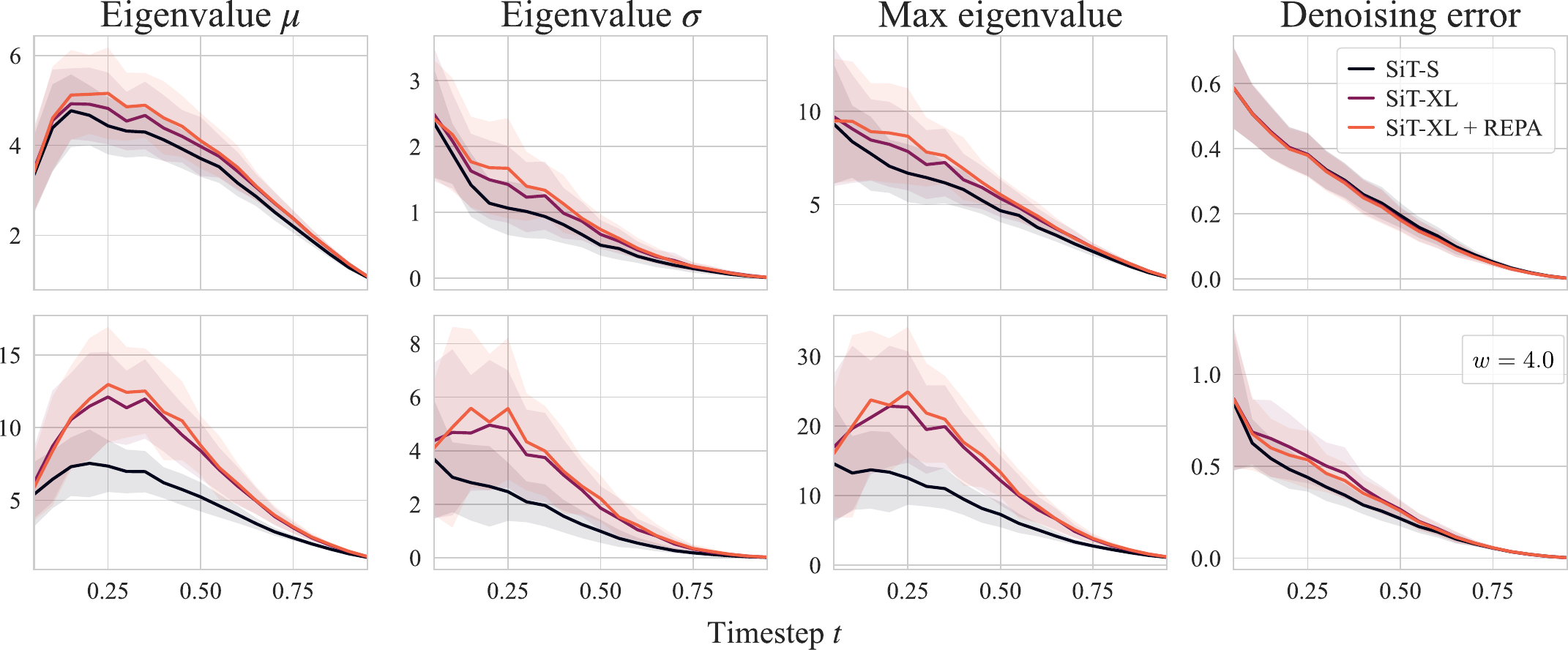}
    \caption{Using Algorithm~\ref{alg:simultaneous_iteration} ($n=10$, $K=10$), we compute eigenvalues of Jacobians for different SiT denoisers. \textbf{Top}: We observe a distinct ordering that aligns with generative performance (SiT-S $<$ SiT-XL $<$ SiT-XL + REPA), whereas denoising capabilities are indistinguishable. \textbf{Bottom}: When using classifer-free guidance with $w=4.0$ the differences between the models are accentuated.}
    \label{fig:measurements}
\end{figure}

\subsection{Spectral Analysis of pre-trained denoiser Jacobians}
\label{sec:analysis}

We analyze the denoiser Jacobian of the state-of-the-art SiT-S, SiT-XL \citep{ma2024sit}, and SiT-XL+REPA \citep{yu2025representation} generative denoising models. The results we obtain are comparable, as all three models were trained on ImageNet \citep{russakovsky2015imagenet} at $256\times256$ resolution and operate in the same VAE latent space \citep{rombach2022high}. We pick $t \in \{0.95, 0.90, \dots, 0.05 \}$, and for each timestep, we randomly sample 100 images from the ImageNet validation set to construct noisy samples $\vx_t$.

We run all models on the same samples, with $\delta=1$, $n=10$ eigenvectors, and $K=10$ iterations for Algorithm~\ref{alg:simultaneous_iteration}. We also re-run the analysis using classifier-free guidance \citep{ho2022classifier}, which is frequently employed to improve sample quality, using guidance scale $w=4$. In Figure~\ref{fig:measurements}, we plot descriptive statistics (mean, standard deviation, and maximum) of the measured eigenvalues, along with the mean squared error between denoised and real images.

We observe that the ordering in eigenvalue statistics ends up being the same as the ordering based on generative performance (FID \citep{heusel2017gans}); SiT-S $<$ SiT-XL $<$ SiT-XL+REPA. The Jacobian of better generative models has larger eigenvalues, notably over the initial and middle denoising steps.
This clearly indicates that larger models not only denoise images more accurately but also use the additional parameters to capture more data variation, as reflected in the wider spectrum of their Jacobian.
Our intuition is that while the smaller SiT-S synthesizes similar-looking images from all inputs in the neighborhood of $\vx_t$, better-performing models produce vastly different results as small perturbations around $\vx_t$ are amplified in the Jacobian. Interestingly, looking only at the denoising error is inconclusive, both with and without guidance.

\begin{figure}[t]
    \centering
    \includegraphics[width=\linewidth]{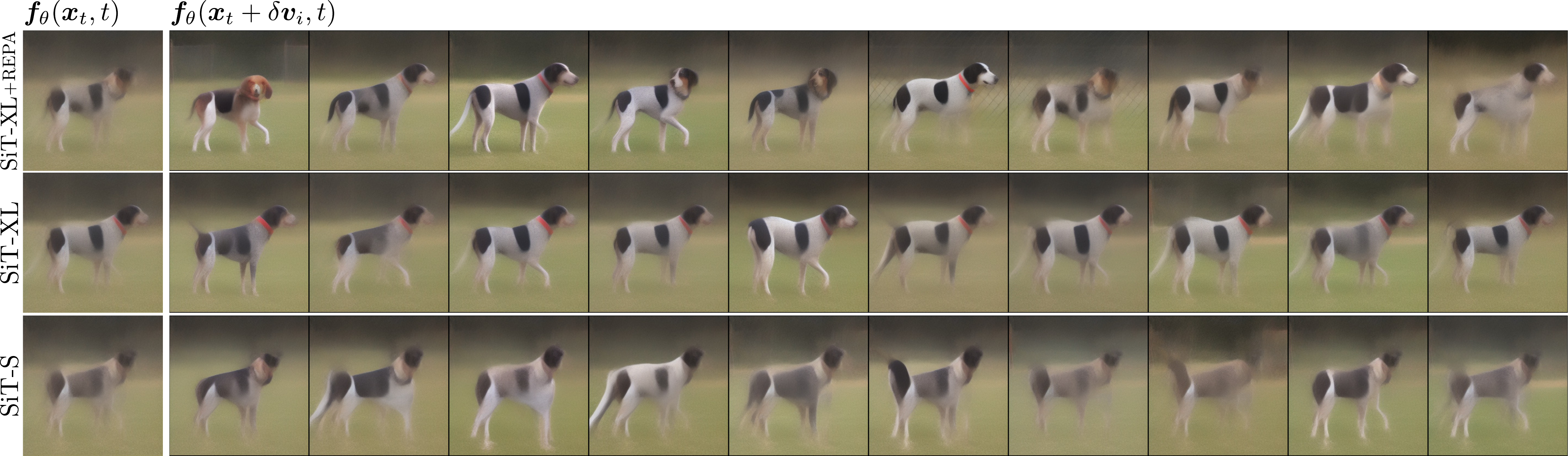}
    \caption{Examples of a denoised image and the eigenvectors computed around it for pre-trained SiT models. We show the top-10 eigenvectors for the same image at $t=0.4$. Better generative models capture more and sharper variability in their top components.}
    \label{fig:eigenvector_examples_ots}
\end{figure}

We visualize eigenvectors across models in Figure~\ref{fig:eigenvector_examples_ots}. Showing $\vv_i$ directly in latent space is uninformative, and instead, we visualize the effect of perturbing the noisy input $\vx_t$ along $\vv_i$, plotting the new prediction $\vf_{\theta}(\vx_t + \delta\vv_i)$.  The new predictions exhibit less variation for the SiT-S eigenvectors, compared to the larger models, which exhibit more meaningful changes. Again, looking only at the denoised image (left) gives no insights about the model's capability as an image generator. We provide additional visualizations of the denoiser Jacobian eigenvectors in Appendix~\ref{sec:viz_eigenvectors}.

\section{Jacobian regularization in training}
\label{sec:jac_regularization}
As shown above, the spectral properties of the denoiser Jacobian constitute strong evidence of the model's generation quality. A question that naturally comes up is whether changing the Jacobian spectrum controls the model's generative capabilities. In this next section, we propose a way to regularize the denoiser training that can control the spectral properties of the learned Jacobian matrix.

We revisit the denoising objective of Eq.~(\ref{eq:denoising}), dropping the weight for simplicity, and add a second term that denoises a slightly perturbed input
\begin{equation}
    \mathcal{L}(\theta) =
    \lVert \underbrace{\vf_{\theta}(\vx_t,t) - \vx_0}_{\vr} \rVert_2^2
    + \tau \left\lVert \vf_{\theta}(\vx_t + \delta\vv,t) - \vx_0 \right\rVert_2^2
    \label{eq:denoising_jac}
\end{equation}
where $\vv$ is a perturbation direction. If $\vr$ is the residual between the real and denoised image that the denoiser minimizes,
the effect of perturbing the input is illustrated with the first-order approximation
\begin{equation}
    \mathcal{L}(\theta) \approx 
    \left\lVert \vr \right\rVert_2^2  + \tau \left\lVert \vr + \delta\mJ_{\theta}\vv \right\rVert_2^2.
    \label{eq:denoising_jac_taylor}
\end{equation}
The perturbed denoising term introduces the Jacobian explicitly in the training objective. Thus, we can regularize the Jacobian in a controlled manner.
Different choices of the perturbation direction $\vv$ regularize the denoiser Jacobian. We discuss these choices below, using the toy 2D mixture-of-Gaussians setting of Figure~\ref{fig:toy} (a) and provide implementation details in Appendix~\ref{sec:toy_experiment}.

\textbf{Stochastic $\vv$}:
For a random $\vv$, with $\mathbb{E}[\vv] = 0$, the expectation of Eq.~(\ref{eq:denoising_jac_taylor}) w.r.t. $\vv$ gives us
\begin{equation}
    \mathbb{E}_{\vv}\left[ \mathcal{L}(\theta) \right] =
    (1+\tau) \left\lVert \vr \right\rVert_2^2 
    + \tau \delta^2 \mathbb{E}_{\vv} \left[ \left\lVert \mJ_{\theta}\vv \right\rVert_2^2 \right].
    \label{eq:rand_objective}
\end{equation}
The objective minimizes both the denoising loss and a Jacobian regularization term that penalizes expansion along the directions of $\vv$. For a Normal $\vv$, this regularization constrains the Jacobian expansion isotropically, reducing to a Frobenius norm regularizer 
$\mathbb{E}_{\vv \sim \normal(\vzero,\mI)}[ \lVert \mJ_{\theta}\vv \rVert_2^2 ] = \lVert \mJ_{\theta} \rVert_F^2$ \citep{scarvelis2024nuclear}.

To illustrate more fine-grained control over the directions in which we allow the Jacobian to expand, we choose $\vv \in \{ \pm [ 0,1]^T, \pm [1,0]^T\}^T$, which minimizes sample variation just along the horizontal and vertical axes. Figure~\ref{fig:toy} (b) shows how a denoiser trained with this regularization uses `squares' to fit each Gaussian component.

\textbf{Residual $\vv$}:
To directly regularize eigenvalues, we would first have to compute them using a power iteration, as in Algorithm~\ref{alg:simultaneous_iteration}. Doing this at every training step is prohibitively slow; we want to design a regularizer that increases the Jacobian response along its eigenvectors without having to compute them.
Looking at Eq.~(\ref{eq:denoising_jac_taylor}), choosing $\vv = -\vr = \vx_0 - \vf_{\theta}(\vx_t,t)$ the objective becomes
\begin{equation}
    \mathcal{L}(\theta) 
    = \left\lVert \vr \right\rVert_2^2 + \tau \left\lVert \vr - \delta\mJ_{\theta}\vr \right\rVert_2^2
    \label{eq:residual_objective}
\end{equation}
which minimizes both the denoising $\lVert \vr \rVert$ and the Jacobian regularization $\lVert \vr-\delta\mJ_{\theta}\vr \rVert$. This regularizer pushes the model to produce a Jacobian response along the direction of the residual, such that $\mJ_{\theta}\vr \approx \frac{1}{\delta}\vr$. The model prediction in $\vr$ is used as a fixed target, and we do not backpropagate through the input perturbation.

With an appropriately small $\delta$, we increase the expansion along the residual, which we expect to increase the overall Jacobian response. We empirically test this hypothesis in Figure~\ref{fig:toy} (c), where we train the denoiser using a fixed $1/\delta=10$. The residual-regularized model exhibits larger eigenvalues in its Jacobian, resulting in fitting the target distribution better, with fewer samples between modes.

\begin{figure}[t]
    \centering
    \includegraphics[width=1.\linewidth]{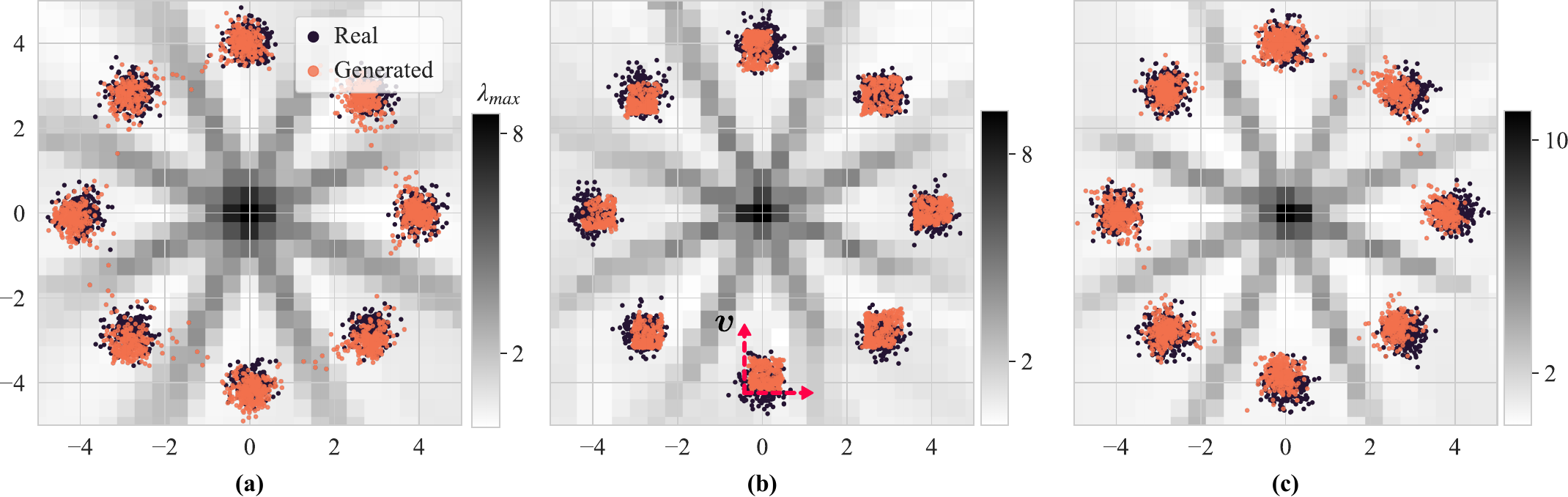}
    \caption{Training a denoising generative model on a mixture of 2D Gaussians. The grayscale color represents the maximum eigenvalue of the Jacobian at $t=0.5$ for each point on the grid. 
    \textbf{(a)} Baseline model trained without regularization. \textbf{(b)} Jacobian regularization using a perturbation that minimizes variation along the orthogonal axes. \textbf{(c)} Jacobian regularization using the residual perturbation, which increases eigenvalues (showing $\lambda_{\max}$), and results in fewer samples falling between modes.}
    \label{fig:toy}
\end{figure}

\textbf{Why does the residual increase eigenvalues?}
The residual direction naturally emerges as an eigenvector-proxy regularization from Eq.~(\ref{eq:denoising_jac_taylor}) by imposing $\mJ_{\theta}\vr \approx \frac{1}{\delta}\vr$, and performs well in the 2D setting of Figure~\ref{fig:toy} (c). On the contrary, naively maximizing the Jacobian response in every direction does not guarantee an increase in the top eigenvalues (Appendix~\ref{sec:contrast_random}) and can instead produce a model sensitive to all perturbations. 

To understand why the residual works well as a perturbation direction for regularization, we consider its covariance
\begin{equation}
    \mathbb{E} [ \vr \vr^T \mid \vx_t ] = 
    \mathbb{E} [ (\vf_{\theta}(\vx_t,t) - \vx_0)(\vf_{\theta}(\vx_t,t) - \vx_0)^T  \mid \vx_t ] = 
    \operatorname{Cov}(\vx_0 \mid \vx_t) = \frac{(1-t)^2}{t} \mJ
    \label{eq:residual_jacobian}
\end{equation}
which directly relates it to the Jacobian through the covariance of the denoiser posterior over $\vx_0$ (Eq.~\ref{eq:jac_covariance}). This allows us to quantify the alignment between the residual and Jacobian eigenvectors. We project the residual onto the eigenvector basis $\vv_i$ of $\mJ$, for a given $\vx_t$, as
\begin{equation}
    \vr = \sum_i \vv_i^T \vr \vv_i = \sum_i p_i \vv_i
\end{equation}
where $p_i$ measures the overlap of the residual $\vr$ with eigenvector $\vv_i$. The residual covariance can be expressed using the eigenvector basis as
\begin{equation}
    \mathbb{E}[ \vr\vr^T \mid \vx_t ] =
    \mathbb{E} \bigl[ \sum_{i,j} p_i p_j \vv_i \vv_j^T \mid \vx_t \bigr]
    =\sum_{i,j} \mathbb{E}[ p_i p_j \mid \vx_t ]\vv_i \vv_j^T.
\end{equation}
Substituting in Eq.~(\ref{eq:residual_jacobian}) and expanding the Jacobian into its eigenvalues $\lambda_i$ and eigenvectors $\vv_i$
\begin{equation}
    \sum_{i,j} \mathbb{E}[ p_i p_j \mid \vx_t ]\vv_i \vv_j^T
    = \frac{(1-t)^2}{t} \sum_i \lambda_i \vv_i\vv_i^T
\end{equation}
where the RHS defines an orthonormal basis over $\mathbb{R}^N$, and we can derive that
\begin{equation}
    \mathbb{E}[ p_i^2 \mid \vx_t  ]= \frac{(1-t)^2}{t}\lambda_i.
    \label{eq:residual_overlap}
\end{equation}
The overlap $p_i$ between the residual direction and an eigenvector is determined by the corresponding eigenvalue $\lambda_i$. Therefore, by increasing the Jacobian response along $\vr$, we increase expansion disproportionately more along the principal eigenvectors, justifying why the residual regularizer produced a larger $\lambda_{\max}$. In contrast, a voluminous increase in the Jacobian spectrum results in an overly sensitive denoiser that fails to learn to sample from the target distribution (Appendix~\ref{sec:contrast_random}).

\section{ImageNet Experiments}
\label{sec:setup_imagenet}

The spectral analysis (Section~\ref{sec:analysis}) showed that in ImageNet-trained SiT models, the Jacobian spectrum is related to generative performance. Next, Section~\ref{sec:jac_regularization} proposed a regularizer that controls the Jacobian spectrum by decreasing or amplifying its response along random and eigen-directions, respectively. We next study whether applying this training regularization to ImageNet SiT models leads to the improvements in generation quality we observed in our initial analysis.

\subsection{Training setup}
We use the SiT-S model and provide further results for SiT-B and UNet in Appendix~\ref{sec:additional_results}. We train the baseline with the $\vx_0$-prediction objective of Eq.~(\ref{eq:denoising}), per-timestep weighting $w(t)=1/(1-t)^2$, and in the latent space of the SD-1.5 VAE \citep{rombach2022high}. We make two necessary modifications to facilitate ImageNet-scale training with the proposed Jacobian regularization.

First, although a fixed $1/\delta$ worked in the toy setting, our analysis of pre-trained models showed that eigenvalues varied significantly across timesteps (Figure~\ref{fig:measurements}). Therefore, we instead opt for an adaptive $1/\delta$, computed using the Rayleigh quotient \citep{horn2012matrix} along the residual direction
$R(\vr) = {\vr^T \mJ_{\theta}\vr}/{\lVert \vr \rVert^2}$.
We keep an exponential moving average of the Jacobian expansion across timesteps $R_t$, updated at every iteration using the finite-difference of Eq.~(\ref{eq:finite_difference}) for $\mJ_{\theta}\vr$. We set a multiplicative gain $1/\delta = kR_t$ during training -- to simplify notation, we will refer to $1/\delta$ as the gain $k$, with larger $1/\delta$ imposing larger Jacobian eigenvalues.

Second, increasing the target gain $1/\delta$ amplifies regularization, but diminishes its gradient, as the perturbed $\vx_t - \delta\vr$ collapses to $\vx_t$. To balance target gain and gradients, we mask the residual to retain only the top-5\% pixels. By masking, the perturbed input explicitly contains less information, avoiding collapse. In Appendix~\ref{sec:full_vs_masked} we discuss how masking does not change the regularization target, but instead strengthens its effect by a factor $\eta>1$ determined by the masking ratio.

For the stochastic regularization, we perturb the input with randomly sampled Normal vectors with unit variance. Stochastic perturbations, compared to residual, have a stationary effect throughout training; for the residual direction, the model itself becomes more sensitive towards eigen-directions as it fits the data better, even without regularization. Therefore, the main tuning parameter for the stochastic regularizer is the weight $\tau$, instead of the radius $1/\delta$. We follow the same recipe as in the residual case, but now fix $1/\delta=2$ and vary $\tau=\{1,0.1\}$. 

We also train a model combining both perturbations where we naively add the two regularization terms to the total loss, using the best-performing hyperparameters from the previous experiments for each. This combination tests whether the effect of each regularizer is unique -- if improvements in generative quality accrue then we can say that each affects a different property of the Jacobian spectrum. We provide more details about the overall training process in Appendix~\ref{sec:training_dynamics}.

\subsection{Jacobian regularization controls spectral properties}
\label{sec:jac_regularization_imagenet}

\begin{figure}[t]
    \centering
    \includegraphics[width=\linewidth]{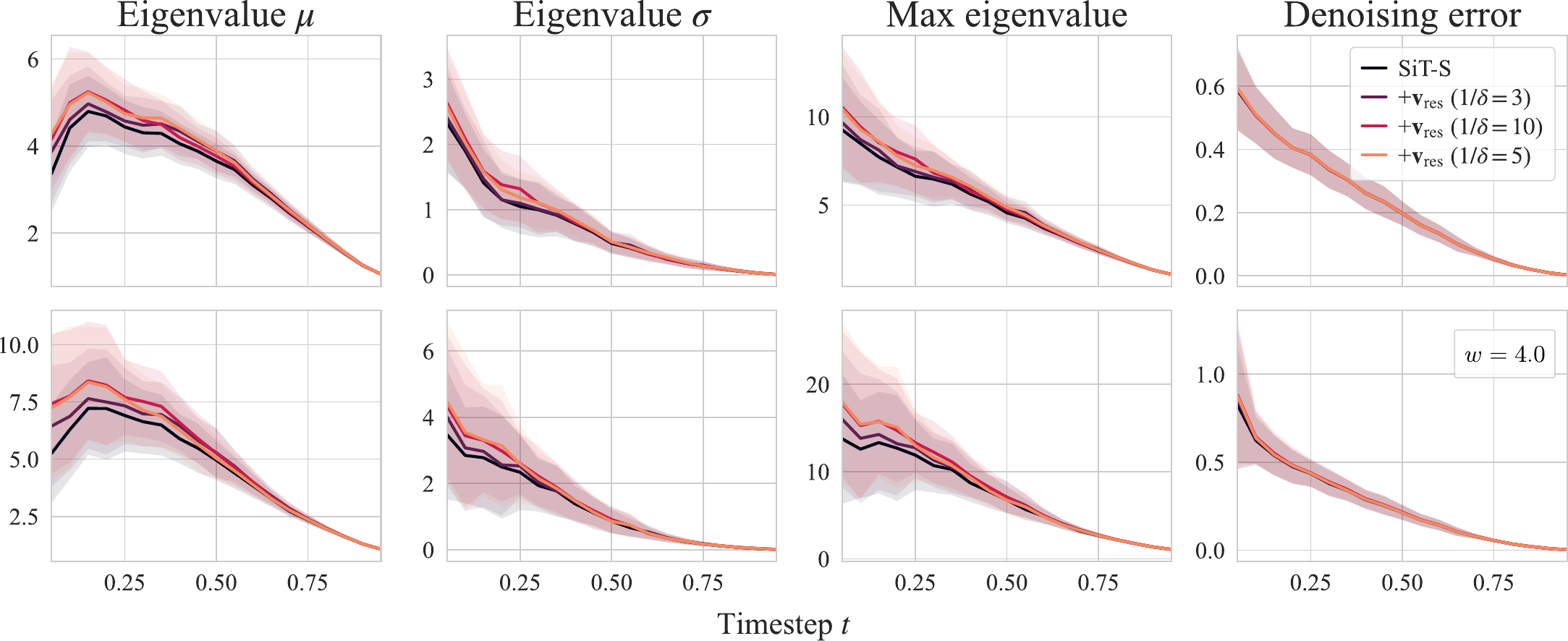}
    \caption{Eigenvalue analysis for SiT-S models trained with residual regularization. By varying the target gain $1/\delta$, we impose larger eigenvalues, with diminishing effects at $1/\delta=10$. Using classifier-free guidance with $w=4.0$ (bottom) amplifies the differences between the models.}
    \label{fig:analysis_sit_s}
\end{figure}
\begin{figure}[t]
    \centering
    \includegraphics[width=\linewidth]{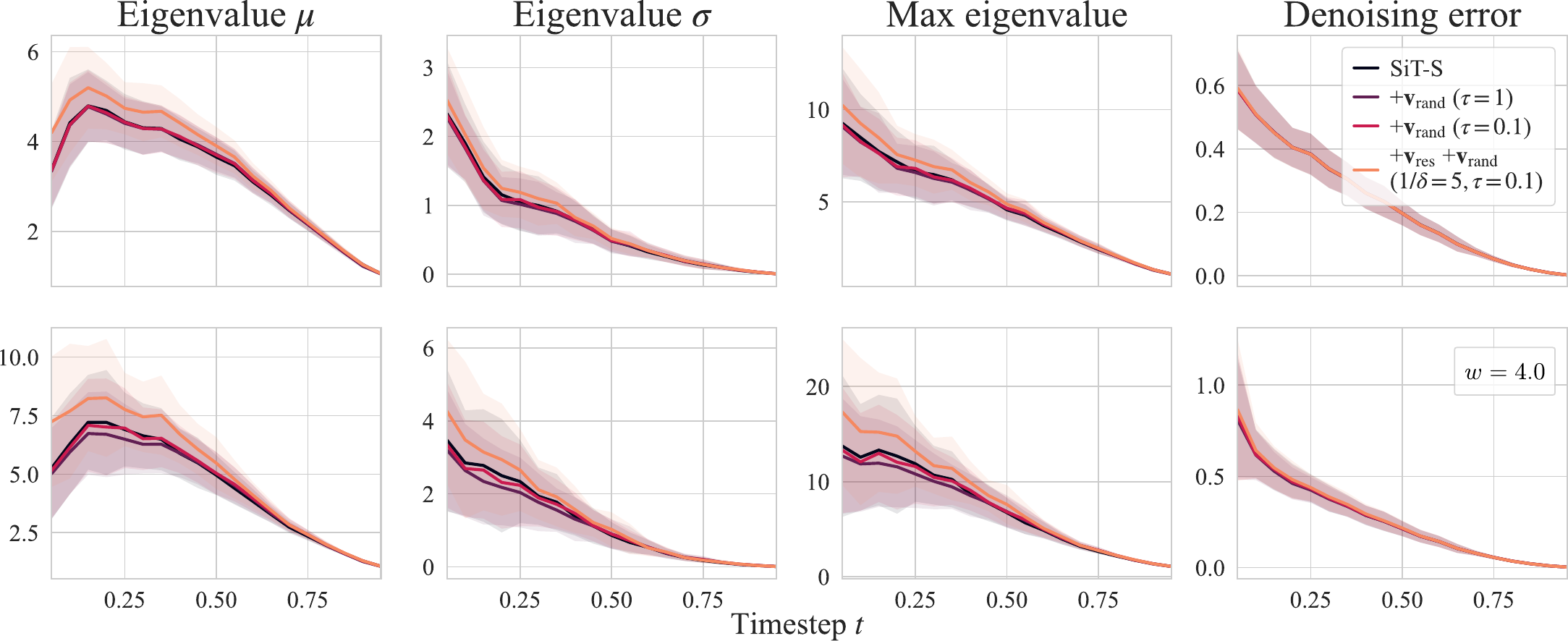}
    \caption{Eigenvalue analysis for SiT-S models trained with stochastic regularization. Using $\tau=0.1$ does not alter the top-10 eigenvalues of the trained model. In contrast, with $\tau=1.0$, the trained model is over-constrained, exhibiting smaller eigenvalues than the baseline model when using guidance with scale $w=4.0$ (bottom). We also include the model combining both regularization signals ($1/\delta=5$, $\tau=0.1$), which we show obtains similar spectra to the residual-only regularized models.}
    \label{fig:analysis_sit_s_rand}
\end{figure}

In Figure~\ref{fig:analysis_sit_s} we repeat the eigenvalue analysis on the baseline and the models trained with residual regularization ($+\mathbf{v}_{\text{res}}$). We observe a clear increase in mean, standard deviation, and maximum eigenvalues over the baseline, which is again more pronounced in earlier timesteps and when using classifier-free guidance. Across the different target gains we set, a larger $1/\delta$ promotes overall larger eigenvalues, but we see diminishing returns when increasing gain from $5 \rightarrow 10$. Despite using the masking operator, which exactly tries to counteract the diminishing effect of decreasing $\delta$, we see that we cannot set an absurdly large target for the eigenvalues.

We then measure the eigenvalues of the stochastic regularization models ($+\mathbf{v}_{\text{rand}}$) in Figure~\ref{fig:analysis_sit_s_rand}. The regularization weight $\tau$ controls the strength of the contraction across all eigenvalues, as it directly weighs the Frobenius penalty $\lVert \mJ_{\theta}\rVert_F$. Classifier-free guidance shows that the $\tau=1$ model is over-constrained, having smaller eigenvalues than the baseline model for its top components. In comparison, a moderate $\tau=0.1$ gives the same spectral statistics as the baseline model, with the stochastic regularization reducing the Jacobian response along random, data-irrelevant directions and not affecting the principal components.

For the model trained with both regularization directions ($+\mathbf{v}_{\text{res}}$ $+\mathbf{v}_{\text{rand}}$ in Figure~\ref{fig:analysis_sit_s_rand}), we choose the best-performing hyperparameters from the previous experiments, $1/\delta=5$ and $\tau=0.1$. The combined model exhibits similar eigenvalue statistics to the residual-only regularized model, which we interpret as the moderate stochastic regularization not affecting the regularization of the top eigen-directions that the residual imposes, but instead suppressing the spurious, data denoising-unrelated components.

\subsection{Jacobian regularization improves generative performance}
\label{sec:generative_performance}

\begin{figure}[t]
    \begin{minipage}[t]{0.39\linewidth}
        \vspace{0pt}
        \centering
        \includegraphics[width=\linewidth]{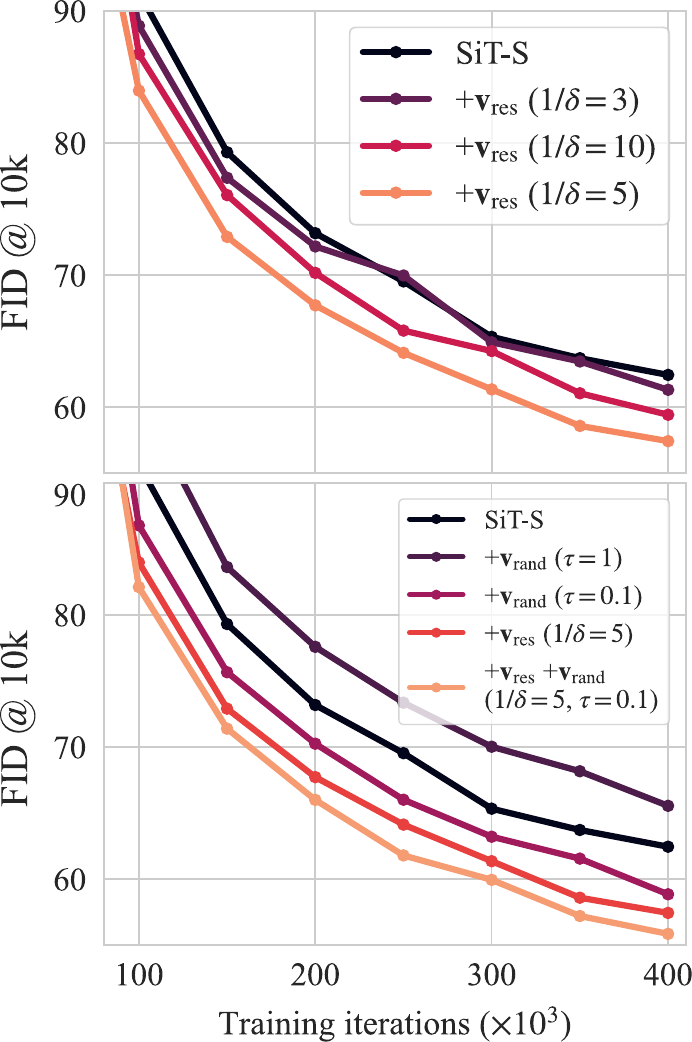}
        \captionof{figure}{FID for the different regularizers and hyperparameters.}
        \label{fig:fid}
    \end{minipage}%
    \hfill
    \begin{minipage}[t]{0.58\linewidth}
        \vspace{0pt}
        \centering
        \captionof{table}{Evaluation metrics varying sampling steps and guidance scales. 
        Each regularization uses the best-performing model 
        ($+\mathbf{v}_{\text{res}}$: $1/\delta=5$), ($+\mathbf{v}_{\text{rand}}$: $\tau=0.1$)
        and 50k images drawn with the Euler-Maruyama sampler.}
        \label{tab:results}
        \resizebox{\linewidth}{!}{\begin{tabular}{lcccc}
    \toprule
    Model & FID \textdownarrow & sFID \textdownarrow & Precision \textuparrow & Recall  \textuparrow \\
    \midrule
    \multicolumn{5}{l}{\small \textit{250 steps, $w=1.0$}} \\[0.25em]
    SiT-S & 59.81 & 9.49 & 0.5929 & 0.3980 \\ 
    $+\mathbf{v}_{\text{res}}$ & 54.43 & 7.88 & 0.6053 & 0.3870 \\ 
    $+\mathbf{v}_{\text{rand}}$ & 56.12 & 8.86 & 0.6027 & 0.4177 \\ 
    $+\mathbf{v}_{\text{res}}$ $+\mathbf{v}_{\text{rand}}$ & 52.43 & 7.79 & 0.6202 & 0.4017 \\
    \arrayrulecolor{gray!60}
    \cmidrule(lr){1-5}
    \arrayrulecolor{black}
    \multicolumn{5}{l}{\small \textit{250 steps, $w=1.5$}} \\[0.25em]
    SiT-S & 30.56 & 6.91 & 0.5339 & 0.5642 \\ 
    $+\mathbf{v}_{\text{res}}$ & 29.19 & 7.86 & 0.5608 & 0.5489 \\
    $+\mathbf{v}_{\text{rand}}$ & 27.73 & 6.69 & 0.5411 & 0.5884 \\
    $+\mathbf{v}_{\text{res}}$ $+\mathbf{v}_{\text{rand}}$ & 27.10 & 7.74 & 0.5530 & 0.5678 \\
    \arrayrulecolor{gray!60}
    \cmidrule(lr){1-5}
    \arrayrulecolor{black}
    \multicolumn{5}{l}{\small \textit{20 steps, $w=4.0$}} \\[0.25em]
    SiT-S & 13.04 & 8.58 & 0.1887 & 0.8465 \\ 
    $+\mathbf{v}_{\text{res}}$ & 12.67 & 8.41 & 0.2062 & 0.8367 \\
    $+\mathbf{v}_{\text{rand}}$ & 12.78 & 8.19 & 0.1748 & 0.8690 \\
    $+\mathbf{v}_{\text{res}}$ $+\mathbf{v}_{\text{rand}}$ & 12.34 & 8.42 & 0.2051 & 0.8524 \\
    \bottomrule
\end{tabular}}
    \end{minipage}%
\end{figure}

The eventual question is whether differences in spectrum are reflected in generative performance. In Figure~\ref{fig:fid} we calculate the FID during training for the SiT-S models, with and without regularization. First, we observe that $+\mathbf{v}_{\text{res}}$ improves FID, and the improvements align with the increase in spectra; the gain resulting in the largest eigenvalues ($1/\delta=5$) also improves FID the most, and diminishing returns ($1/\delta=10$) perform worse due to over-regularization. For the stochastic $+\mathbf{v}_{\text{rand}}$ regularization, $\tau=1$ is worse than the baseline, penalizing useful eigenvectors and hurting generative performance. The more moderate $\tau=0.1$, for which the top end of the spectrum did not change, yields an improved FID score between the baseline and the residual-regularized model.

Additionally, we test the baseline and best-performing regularized models in different settings in Table~\ref{tab:results}. The $+\mathbf{v}_{\text{res}}$ model benefits less from classifier-free guidance, showing smaller improvements than other models. We attribute this to increased sensitivity, amplified by classifier-free guidance, that can lead to larger error accumulation during sampling. This difference is notably reduced for fewer steps (20), hinting at sampling steps as the culprit. Finally, we observe small variations in fidelity (FID, precision) and diversity (recall) metrics among the models.

We close this discussion by considering the model that combines both regularization objectives. The model achieves the best overall FID, suggesting that the regularizers act on complementary properties of the Jacobian spectrum. We interpret this as follows: the denoising objective identifies Jacobian directions from the data that are useful in reconstructing clean samples. The residual Jacobian regularizer further strengthens these task-relevant directions, while the stochastic regularizer attenuates weak directions that correspond to off-manifold or spurious variations. Our results suggest that the effect of regularization is spread across the entire Jacobian spectrum, with suppression of weak spurious components and amplification of useful, data-aligned directions.

\section{Limitations}
\label{sec:limitations}
The residual is constructed from the initial model prediction, and thus requires a second forward pass through the denoiser. This considerably slows training compared to other regularization methods, which can usually be applied in a single forward pass. Furthermore, the increased sensitivity induced by the residual regularization weakens the benefits of classifier-free guidance. Although resolved when combining with the random perturbation, a better approach could make the regularizer timestep-dependent and only increase eigenvalues at timesteps where they are most beneficial.

\section{Conclusion}
\label{sec:conclusion}

In this work, we showed that the Jacobian of the denoisers used in diffusion and flow-matching models encodes useful information about their generative capabilities. Using spectral analysis, we showed how models that performed better in synthesizing samples from the learned distributions also have larger principal Jacobian eigenvalues. This observation motivated a regularization objective to control the Jacobian spectrum during training. This allowed us to test whether the relationship is bidirectional, whereby changing these spectral properties resulted in improved generation quality. Our findings have two main implications. First, we establish that the denoiser Jacobian provides a useful tool for identifying differences between models, with the proposed spectral analysis being just one of the ways to probe it. Second, identifying such differences informs the design of the training process itself, allowing us to impose the desirable denoiser properties in training, improving efficiency and performance of the model itself.


\bibliography{main}
\bibliographystyle{iclr2027_conference}

\clearpage
\appendix
\section{Jacobian and covariance}
\label{sec:jac_covariance}

The denoiser Jacobian is related to the covariance matrix of the learned denoising distribution:
Let $\vx_0$ denote the clean signal and let $\vx_t$ be its noisy observation under additive Gaussian noise:
\begin{equation}
    \vx_t = t\vx_0 + (1-t)\vepsilon,\quad \vepsilon \sim \mathcal{N}(\vzero,\mI).
\end{equation}
By Bayes' rule,
\begin{equation}
    p(\vx_0 \mid \vx_t) = \frac{p(\vx_t \mid \vx_0)p(\vx_0)}{p(\vx_t)},\ \text{where}\ 
    p(\vx_t \mid \vx_0) \propto \exp\left(-\frac{\lVert \vx_t-t\vx_0\rVert_2^2}{2(1-t)^2}\right).
\end{equation}
Taking the logarithm of the posterior
\begin{equation}
    \log p(\vx_0 \mid \vx_t) = \log p(\vx_t \mid \vx_0) + \log p(\vx_0) - \log p(\vx_t)
\end{equation}
which, differentiating w.r.t. $\vx_t$, gives
\begin{equation}
    \nabla_{\vx_t} \log p(\vx_0 \mid \vx_t) = \nabla_{\vx_t} \log p(\vx_t \mid \vx_0) 
    - \nabla_{\vx_t} \log p(\vx_t)
    = \frac{t\vx_0-\vx_t}{(1-t)^2} - \nabla_{\vx_t} \log p(\vx_t).
    \label{eq:posterior-score}
\end{equation}
Tweedie's formula \citep{efron2011tweedie} states that
\begin{equation}
    \underbrace{\mathbb{E}\left[ \vx_0 \mid \vx_t \right]}_{\hat{\vx}_0} 
    = \frac{1}{t}\vx_t + \frac{(1-t)^2}{t} \nabla_{\vx_t} \log p(\vx_t).
\end{equation}
or equivalently
\begin{equation}
    \nabla_{\vx_t} \log p(\vx_t) = \frac{t\hat{\vx}_0-\vx_t}{(1-t)^2}.
\end{equation}
Substituting this expression into Eq.~(\ref{eq:posterior-score})
\begin{equation}
    \nabla_{\vx_t} \log p(\vx_0 \mid \vx_t) = \frac{t\vx_0-\vx_t}{(1-t)^2}
    - \frac{t\hat{\vx}_0-\vx_t}{(1-t)^2} = \frac{t(\vx_0-\hat{\vx}_0)}{(1-t)^2}
\end{equation}
which we can write as
\begin{equation}
    \frac{\nabla_{\vx_t} p(\vx_0 \mid \vx_t)}{p(\vx_0 \mid \vx_t)} 
    = \frac{t(\vx_0-\hat{\vx}_0)}{(1-t)^2},
\end{equation}
and hence
\begin{equation}
    \nabla_{\vx_t} p(\vx_0 \mid \vx_t) = 
    \frac{t(\vx_0-\hat{\vx}_0)}{(1-t)^2} p(\vx_0 \mid \vx_t).
\end{equation}
Now we can consider the Jacobian of the posterior mean:
\begin{align}
    \mJ &= \frac{\partial \hat{\vx}_0}{\partial \vx_t}
    = \frac{\partial}{\partial \vx_t} \mathbb{E}\left[ \vx_0 \mid \vx_t \right]
    = \frac{\partial}{\partial \vx_t} \int \vx_0\,p(\vx_0 \mid \vx_t)\,d\vx_0 \notag\\
    &= \int \vx_0 \left[ \nabla_{\vx_t} p(\vx_0 \mid \vx_t) \right]^T d\vx_0
    = \frac{t}{(1-t)^2} \int \vx_0 (\vx_0-\hat{\vx}_0)^T p(\vx_0 \mid \vx_t)\,d\vx_0 \notag\\
    &= \frac{t}{(1-t)^2} \left( \int \vx_0\vx_0^T p(\vx_0 \mid \vx_t)\,d\vx_0
    - \int \vx_0\hat{\vx}_0^T p(\vx_0 \mid \vx_t)\,d\vx_0 \right) \notag\\
    &= \frac{t}{(1-t)^2} \left( \mathbb{E}\left[ \vx_0\vx_0^T \mid \vx_t \right]
    - \hat{\vx}_0 \hat{\vx}_0^T \right)  \notag\\
    &= \frac{t}{(1-t)^2} \left( \mathbb{E}\left[ \vx_0\vx_0^{\top}\mid \vx_t \right]
    - \mathbb{E}\left[ \vx_0 \mid \vx_t \right] 
    \mathbb{E}\left[ \vx_0 \mid \vx_t \right]^T \right) \notag\\
    &= \frac{t}{(1-t)^2} \operatorname{Cov}(\vx_0 \mid \vx_t).
\end{align}
This connects the Jacobian of an MMSE-optimal denoiser to the posterior covariance learned by the model
\begin{equation}
    \boxed{\mJ_{\theta} = \frac{t}{(1-t)^2} \operatorname{Cov}(\vx_0 \mid \vx_t)}.
\end{equation}

We can also differentiate Tweedie's formula directly, 
\begin{equation}
     \mJ = \frac{1}{t}\mI + \frac{(1-t)^2}{t} \nabla_{\vx_t}^2 \log p(\vx_t)
\end{equation}
which only tells us that the Jacobian matrix is symmetric, as the sum of $\mI$ and a Hessian matrix. Finally, covariance matrices are always positive semi-definite, and thus we expect all eigenvalues of the (ideal) Jacobian matrix to be real and $\lambda_i \geq 0$.

\section{Additional results}
\label{sec:additional_results}

\subsection{Spectral analysis}
\begin{figure}[h]
    \centering
    \includegraphics[width=\linewidth]{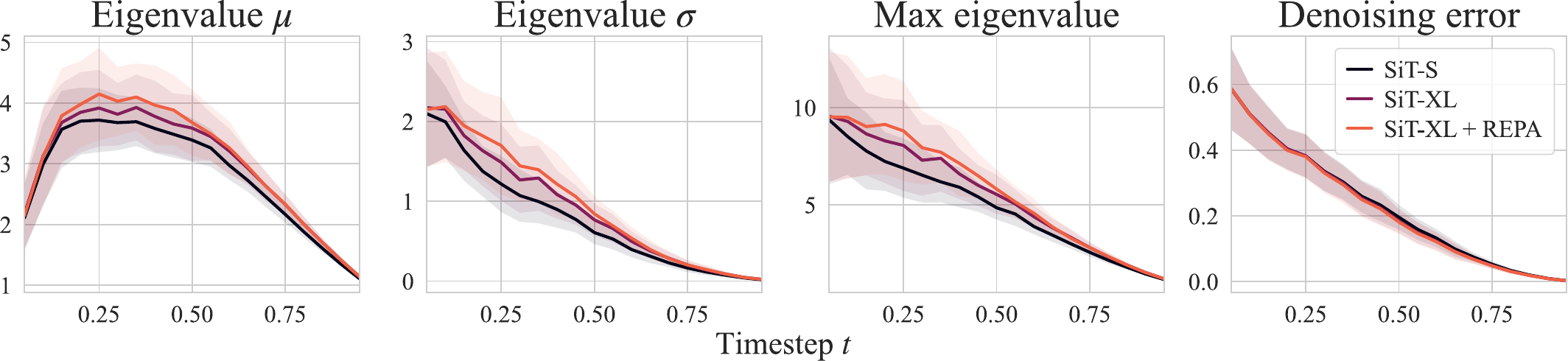}
    \caption{Using Algorithm~\ref{alg:simultaneous_iteration}, we measure eigenvalues of the Jacobians of different flow-matching denoisers. We find an ordering that correlates with the model's expected performance (SiT-S $<$ SiT-XL $<$ SiT-XL + REPA). \textbf{Top}: Using $n=20$ and $K=20$. \textbf{Bottom}: Using classifier-free guidance with scale $w=4.0$.}
    \label{fig:measurements_n20_cfg}
\end{figure}

We extend the analysis of pre-trained models by increasing the number of eigenvectors to $n=20$ and using $K=20$ iterations. The results, shown in Figure~\ref{fig:measurements_n20_cfg}, retain the same ordering but with lower overall values. This is due to the exponentially decaying nature of the eigenvalues, where most of the energy is contained in the first few components.

\begin{wrapfigure}[12]{r}{0.43\linewidth}
    \small
    \vspace{-0.3in}
    \includegraphics[width=\linewidth]{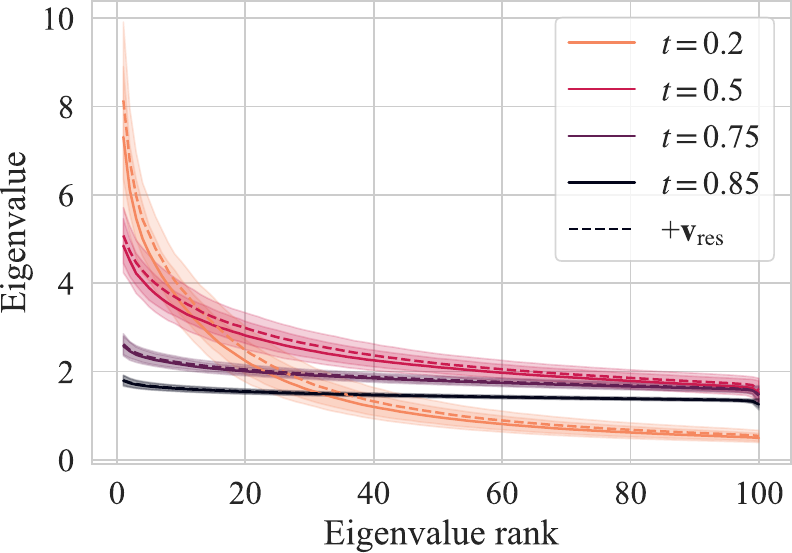}
    \captionsetup{font=small}
    \caption{Comparing the spectrum of the top-100 eigenvalues of SiT-S (solid) and SiT-S with residual regularization (dashed).}
    \label{fig:eigval_spectrum}
\end{wrapfigure}

We plot the spectrum of the top-100 eigenvalues for the trained SiT-S and SiT-S + Jacobian residual regularization models in Figure~\ref{fig:eigval_spectrum}. We observe that the eigenvalues decay exponentially, which justifies limiting our analysis to the top-10 eigenvectors of each model in the main text. Furthermore, the increase in eigenvalues between the baseline and residual-regularized model is focused on the higher end of the spectrum, throughout all noise levels. This supports our analysis of the effect of the proposed residual direction, which we showed focuses on increasing the eigenvalues of directions with largest variance.

\subsection{Extension to other models}
We test how the residual regularization scales with parameters by applying it to an SiT-B model \citep{ma2024sit}. We also apply the residual Jacobian regularization to a UNet-based model \citep{ronneberger2015u} to test whether the regularizer transfers to architectures beyond transformers. We utilize the UNet denoiser configuration from ADM \citep{nichol2021improved} and set up a 30M-parameter model, comparable to the 33M parameters of the SiT-S model used in the main experiments. We again utilize the same VAE latent space as with SiT-S, training baseline and residual-regularized models with different gain targets.

\begin{figure}[t]
    \centering
    \includegraphics[width=\linewidth,trim={0 1.9em 0 0},clip]{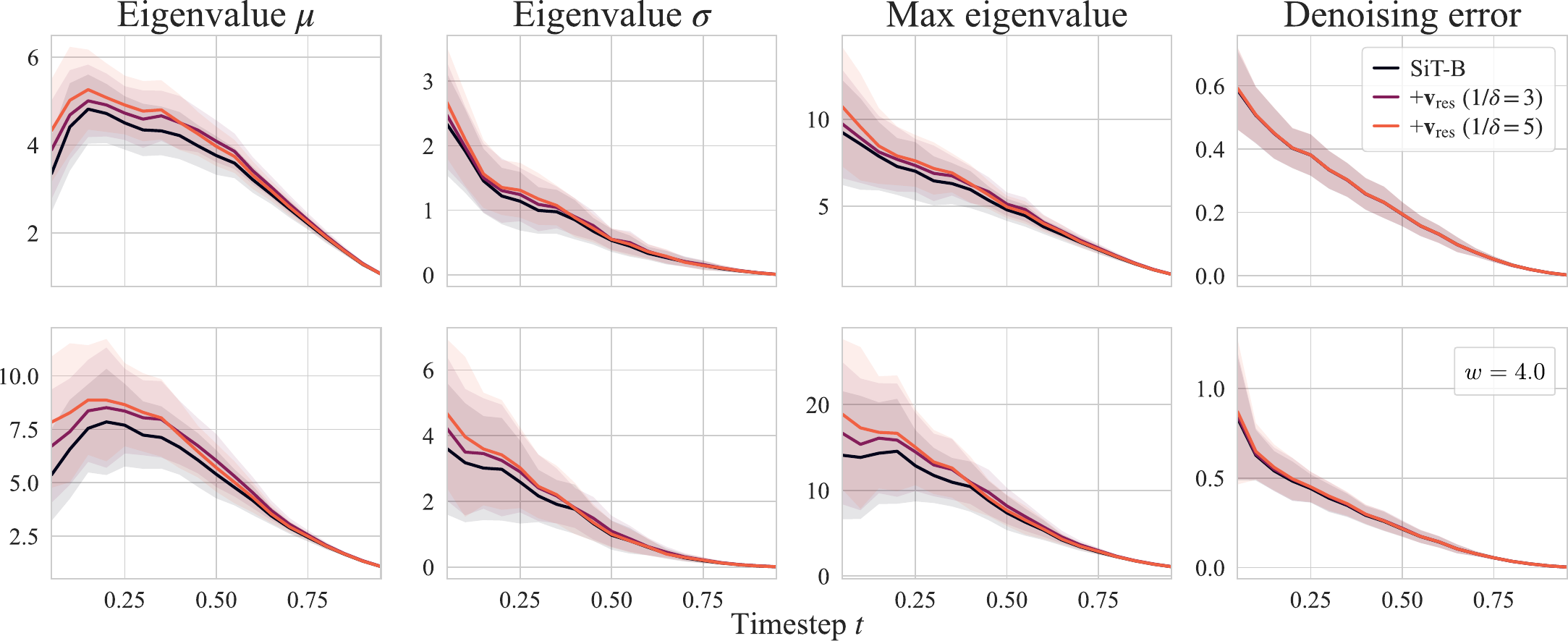}
    \\ \vspace{0.5em}
    \includegraphics[width=\linewidth,trim={0 0 0 2.25em},clip]{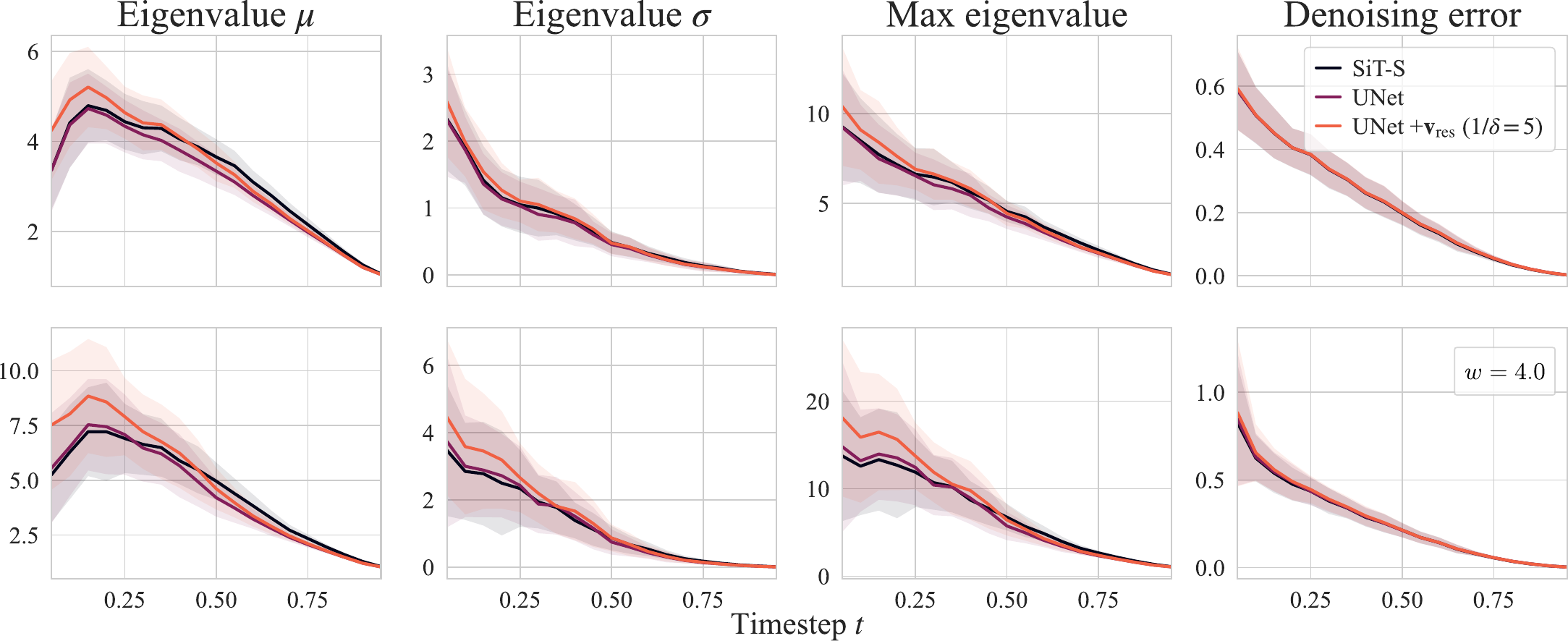}
    \caption{SiT-B and UNet denoiser eigenvalue comparison between baseline and models trained with the proposed residual regularization.}
    \label{fig:analysis_sit_b_unet}
\end{figure}

When we run the eigenvalue analysis, we find that the same observations hold under both a larger model and a different denoiser architecture (Figure~\ref{fig:analysis_sit_b_unet}). The bigger SiT-B model saturates faster; when we increase the gain $1/\delta$ from 3\textrightarrow5, we observe the same diminishing returns we observed for SiT-S when increasing from 5\textrightarrow10. The UNet behaves similarly to the SiT-S, exhibiting comparable eigenvalue statistics.

\begin{figure}[t]
    \begin{minipage}[t]{0.40\linewidth}
        \vspace{0pt}
        \centering
        \includegraphics[width=\linewidth]{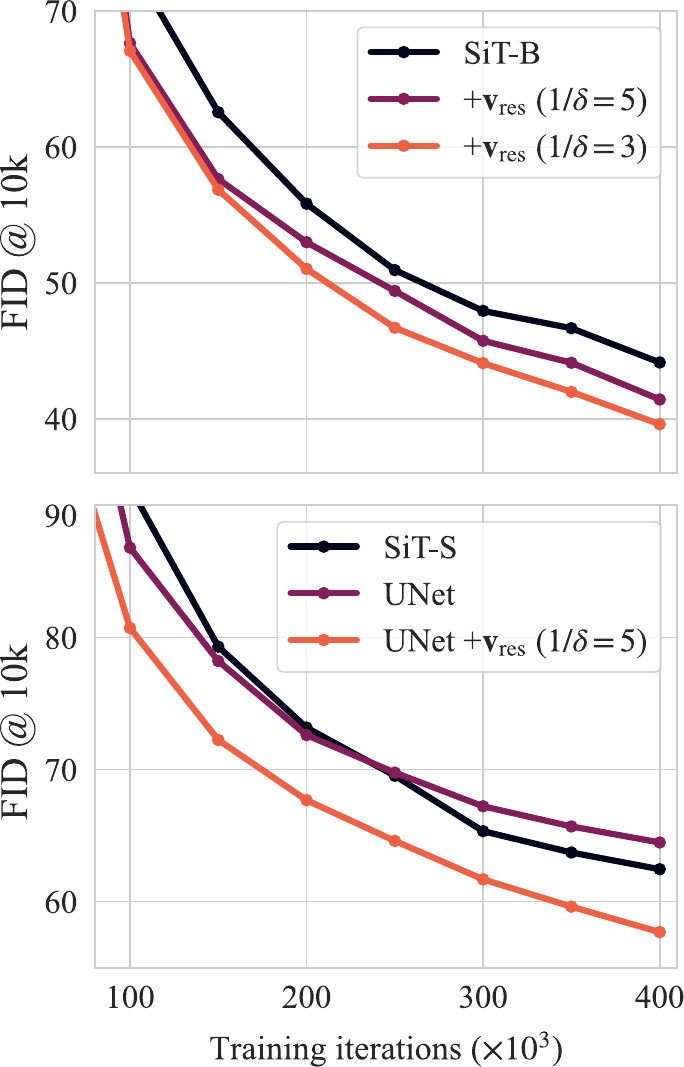}
        \captionof{figure}{FID comparisons for an SiT-B and a UNet-based denoiser trained with and without Jacobian regularization. For the UNet, we include the baseline SiT-S FID as a reference curve.}
        \label{fig:fid_sitb_unet}
    \end{minipage}%
    \hfill
    \begin{minipage}[t]{0.57\linewidth}
        \vspace{0pt}
        \centering
        \captionof{table}{FID results for all trained models. We evaluate the EMA checkpoint of each model at 400k training iterations with 50,000 generated samples. We use the Euler-Maruyama sampler, 250 inference steps, and no classifier-free guidance. The two hyperparameters we control are the residual gain target ($1/\delta$) and the regularization weight for the random perturbations ($\tau$).}
        \label{tab:fid_all}
        \begin{tabular}{lc}
    \toprule
    Model & FID @ 50k \textdownarrow \\
    \midrule
    SiT-S                         & 59.78 \\[0.25em]
    $+\mathbf{v}_{\text{res}}$ ($1/\delta = 2$)  & 59.86 \\
    $+\mathbf{v}_{\text{res}}$ ($1/\delta = 3$)  & 58.88 \\
    $+\mathbf{v}_{\text{res}}$ ($1/\delta = 5$)  & 54.43 \\
    $+\mathbf{v}_{\text{res}}$ ($1/\delta = 10$) & 56.75 \\
    $+\mathbf{v}_{\text{res}}$ ($1/\delta = 5$, Full) & 59.14 \\
    \arrayrulecolor{gray!60}
    \cmidrule(lr){1-2}
    \arrayrulecolor{black}
    $+\mathbf{v}_{\text{rand}}$ ($\tau=1$)   & 63.19 \\
    $+\mathbf{v}_{\text{rand}}$ ($\tau=0.1$) & 56.07 \\
    $+\mathbf{v}_{\text{rand}}$ $+\mathbf{v}_{\text{res}}$ ($\tau=0.1$, $1/\delta = 5)$ & 52.43 \\
    \midrule
    SiT-B & 41.12 \\[0.25em]
    $+\mathbf{v}_{\text{res}}$ ($1/\delta = 3)$  & 36.82 \\
    $+\mathbf{v}_{\text{res}}$ ($1/\delta = 5)$  & 38.68 \\
    \midrule
    UNet & 60.76 \\[0.25em]
    $+\mathbf{v}_{\text{res}}$ ($1/\delta = 5)$  & 55.10 \\
    \bottomrule
\end{tabular} 
    \end{minipage}%
\end{figure}

\begin{figure}[t]
    \centering
    \includegraphics[width=1.\linewidth]{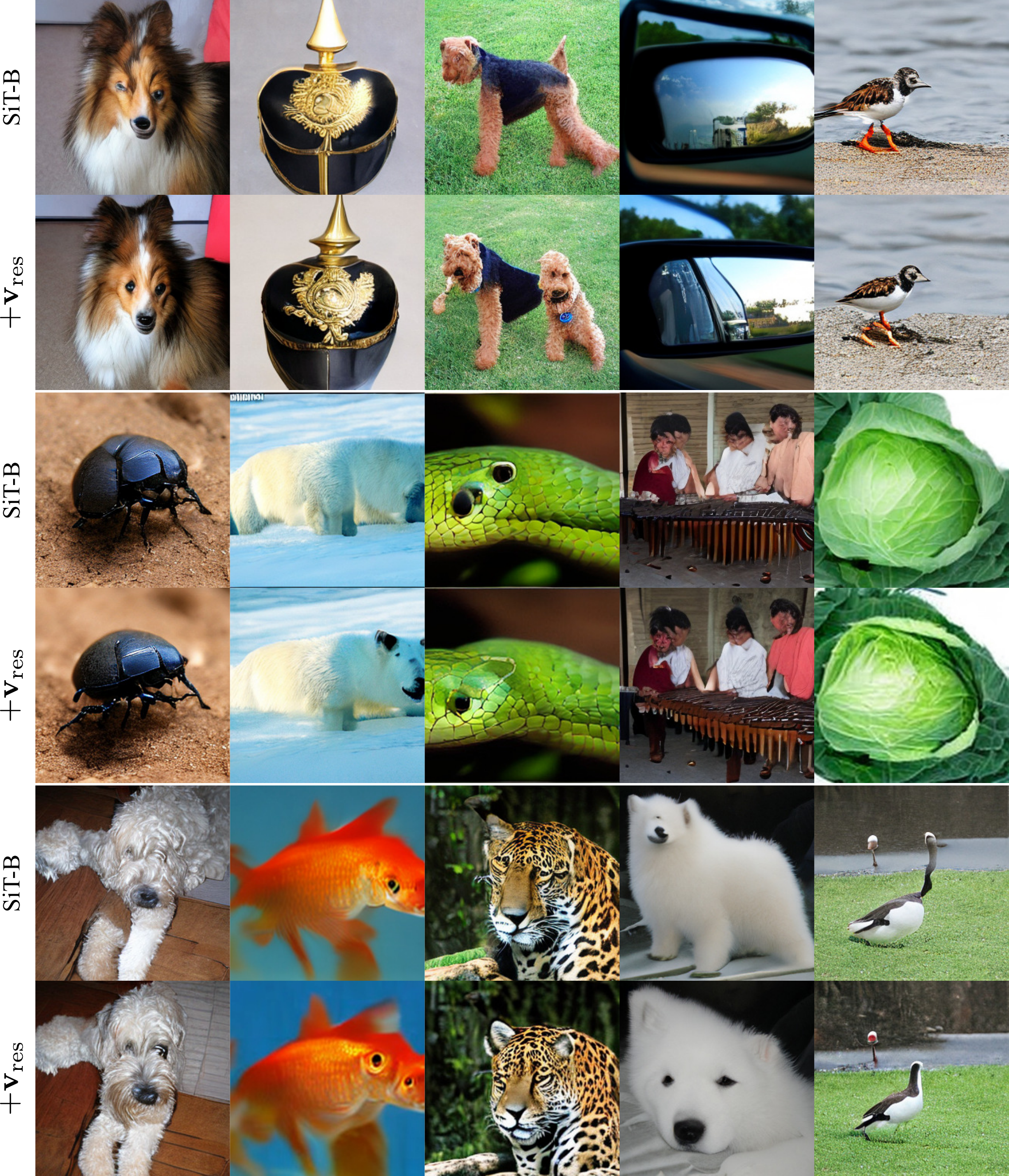}
    \caption{Examples of images generated with the baseline and Jacobian-regularized SiT-B models. We use the Euler sampler with 50 inference steps and guidance scale $w=4.0$.}
    \label{fig:gen}
\end{figure}

We also evaluate the FID of the trained SiT-B and UNet denoisers in Figure~\ref{fig:fid_sitb_unet}. We observe similar improvements over the baseline as in the SiT-S model, with the FID score improving by $\sim$5 points between the baseline and the best-performing regularized model. As with the higher-gain case for the SiT-S model, the SiT-B model with $1/\delta = 5$ performs worse than the less-rigid $1/\delta=3$, which suggests that the optimal $1/\delta$ choice depends on the model and not on the dataset. We provide examples of images generated with the SiT-B and SiT-B + Jacobian regularization models in Figure~\ref{fig:gen}.

We present all FID results in Table~\ref{tab:fid_all} to provide a complete picture of the evaluation. All models were evaluated with 50,000 images using the Euler-Maruyama sampler, 250 inference steps and no classifier-free guidance. We compare the statistics of the generated images to the ImageNet statistics published by ADM \citep{nichol2021improved}. The findings are consistent across different models, with the residual regularization improving fidelity. Additionally, for the SiT-S model, we include the stochastic regularization and combined model results. We did not train a combined regularization variant for the SiT-B and UNet, but expect the performance improvements to be comparable.

\subsection{Comparing residual and Stochastic regularization}

\begin{figure}[t]
    \centering
    \includegraphics[width=1.0\linewidth]{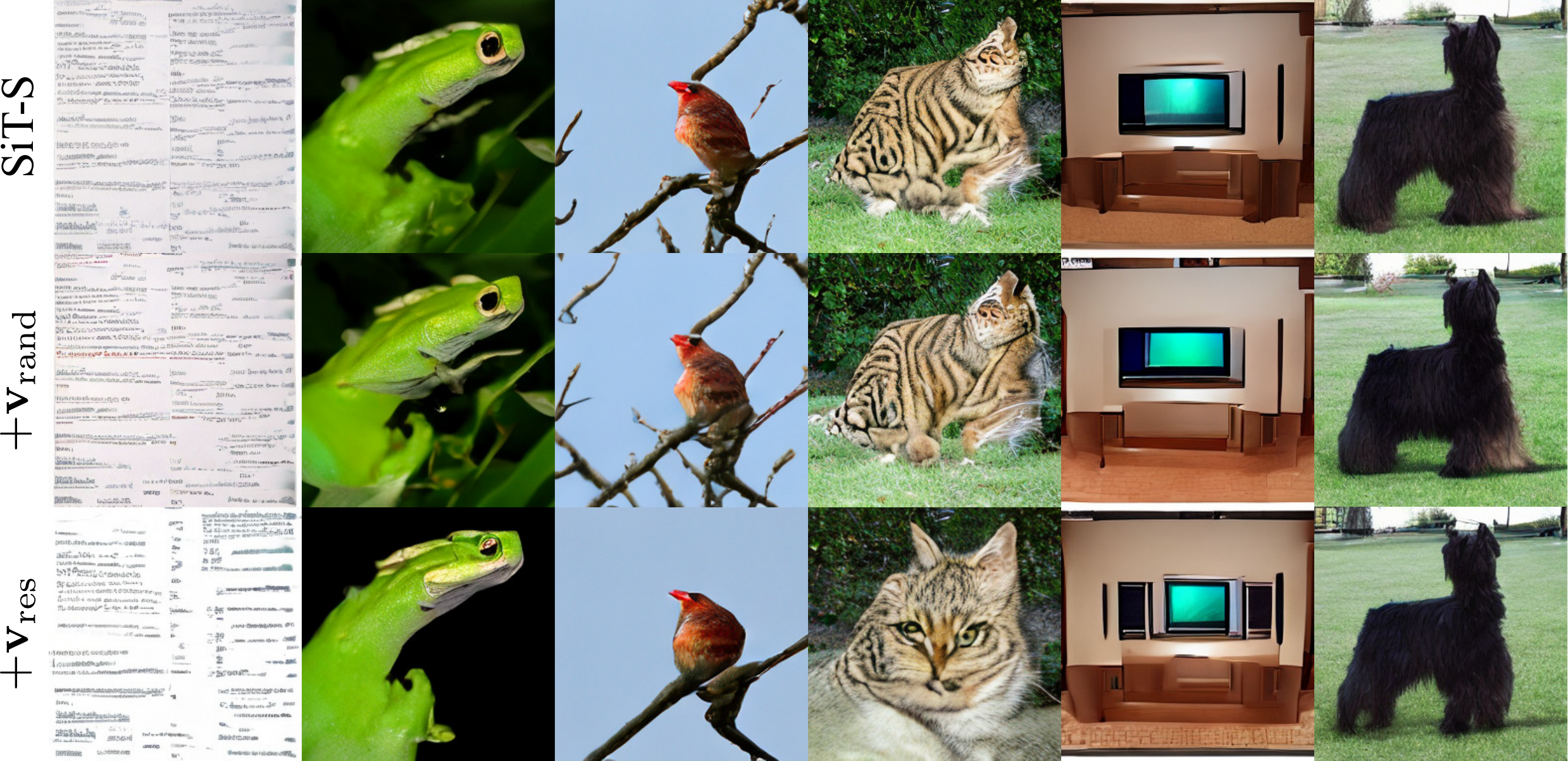}
    \caption{We synthesize images from the same noise, using 50 steps and classifier-free guidance scale $w=4.0$ to amplify differences. We observe that the model using the residual perturbation
    performs larger corrections to the baseline-generated images, indicating a stronger overall effect.}
    \label{fig:masked_vs_rand_gen}
\end{figure}

The stochastic regularized model with $\tau=0.1$ considerably improves over the baseline, while simultaneously not showing a different Jacobian spectrum, as measured in Figure~\ref{fig:analysis_sit_s_rand}. In Figure~\ref{fig:masked_vs_rand_gen} we qualitatively compare generated images from the baseline SiT-S and the stochastic or residual-regularized models. The residual perturbation induces larger corrections on the baseline-generated images, which indicates that this regularization has a stronger effect than random. Despite improving FID, this has also potentially harmful effects, as over-correction can make images unrealistic. We believe that this is the effect that leads to smaller improvements with classifier-free guidance (Section~\ref{sec:generative_performance} and Table~\ref{tab:results}). 

\clearpage
\section{Residual and masking}
\label{sec:full_vs_masked}

\subsection{\texorpdfstring{Effect of $\delta$}{Effect of delta}}
In the main text, we discussed how a masking operator on the residual increases the `push' on the eigenvalue increase without increasing the weight $\tau$ of the regularization term or reducing the step size $\delta$. Regarding $\tau$, we initially experimented with higher values but found no significant effect on the model. To understand the effect of masking the residual, we will first discuss the effect of $\delta$.

Using the regularization term with $\vv = -\vr$ as the perturbation direction, the loss we minimize is
\begin{equation}
    \mathcal{L}(\theta) = \left\lVert \vr \right\rVert_2^2
    + \tau\underbrace{\left\lVert \vr - \delta\mJ\vr \right\rVert_2^2}_{\mathcal{L}_{\text{reg}}}.
\end{equation}
Focusing on the regularization, we can first project the residual onto the eigenvectors of the Jacobian, as $\vr = \sum_i p_i\vv_i$, and then rewrite the regularizer
\begin{equation}
    \mathcal{L}_{\text{reg}}(\theta) = 
    \bigl\lVert \sum_i p_i \vv_i - \delta\mJ \sum_i p_i \vv_i \bigr\rVert_2^2 = 
    \bigl\lVert \sum_i p_i (1 - \delta \lambda_i) \vv_i \bigr\rVert_2^2 = 
    \sum_i p_i^2 (1 - \delta \lambda_i)^2
\end{equation}
Taking the gradient of the regularization term w.r.t. an eigenvalue $\lambda_i$ gives us
\begin{equation}
    \frac{\partial \mathcal{L}_{\text{reg}}}{\partial \lambda_i} = 
    -2 p_i^2 \delta (1 - \delta \lambda_i)
\end{equation}
which has two competing factors. By decreasing $\delta$, we push $\lambda_i$ towards a higher value through $\lambda_i = 1/\delta$, but simultaneously we attenuate the effect of the regularization term since the gradient is multiplied by $\delta$ itself. Therefore, it is impractical to pursue higher eigenvalues in the model only by making delta smaller. By introducing masking, we find a workaround that achieves the desired effect without diminishing the effect of the regularization term.

\begin{figure}[t]
    \centering
    \includegraphics[width=\linewidth]{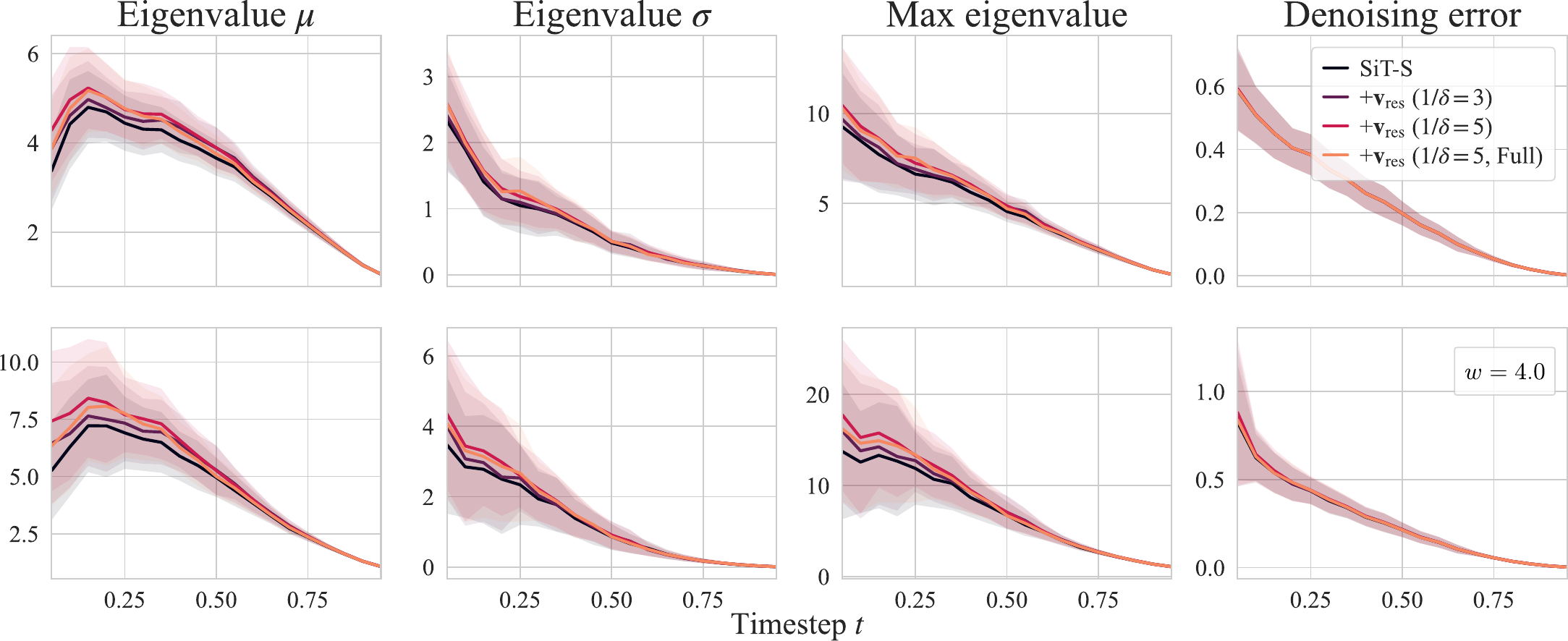}
    \caption{We compare trained model eigenvalues when using the masked and full residual as the perturbation direction. \textbf{Top}: Comparison for $n=10$, $K=10$. \textbf{Bottom}: Same comparison using classifier-free guidance with scale $w=4.0$.}
    \label{fig:analysis_sit_s_full}
\end{figure}

To formulate this, we consider masking as the linear operator $\mM$, which makes the loss of Eq~\ref{eq:residual_objective}
\begin{equation}
    \mathcal{L}(\theta)
    = \lVert \vr \rVert_2^2 + \tau \lVert \vr - \delta\mJ\mM\vr \rVert_2^2.
\end{equation} 
What the regularization term does is push the residual towards
\begin{equation}
    \mJ\mM\vr = \frac{1}{\delta}\vr.
\end{equation} 
The minimum rank positive semi-definite $\mJ$ that satisfies this \citep{horn2012matrix} is
\begin{equation}
    \mJ_{\min} = \frac{1}{\delta\vr^T\mM\vr} \vr\vr^T = \frac{1}{\delta\lVert \mM\vr\rVert^2} \vr\vr^T
\end{equation}
and applying this to the full residual would give
\begin{equation}
    \mJ_{\min}\vr  = \frac{1}{\delta}\frac{\lVert \vr \rVert^2}{\lVert \mM\vr \rVert^2} \vr
    = \frac{1}{\delta}\eta\vr,
    \quad \text{where}\ \eta = \frac{\lVert r \rVert^2}{\lVert \mM\vr \rVert^2} > 1.
    \label{eq:masked_eta}
\end{equation} 
By masking the residual, we still apply the same regularization, increasing the top eigenvalues of the denoiser Jacobian. The difference is that the target gain $1/\delta$ is also multiplied by a factor $\eta > 1$, where $\eta$ depends on the masking ratio, with more masking leading to larger $\eta$. For our ImageNet experiments, we choose a mask that retains the top-5\% residual pixels by average intensity. We did not ablate the masking effect, but found that this extreme masking operation led to good results, as shown in Table~\ref{tab:results}.

\subsection{Training without masking}
We train an SiT-S using the full residual and the best-performing $1/\delta=5$. In Figure~\ref{fig:analysis_sit_s_full} we compare the eigenvalues from different models trained with masking and the non-masked model. The first observation we make is that for the same radius $1/\delta=5$, the masked model Jacobian has larger eigenvalues, especially for lower timesteps, where the full residual has similar mean eigenvalues to a masked model using $1/\delta=3$. This observation is consistent with the result of Eq.~\ref{eq:masked_eta}, which predicts that for the same $1/\delta$, the masked residual will produce higher eigenvalues.

\begin{figure}[t]
    \begin{minipage}[t]{0.42\linewidth}
        \centering
        \includegraphics[width=0.95\linewidth]{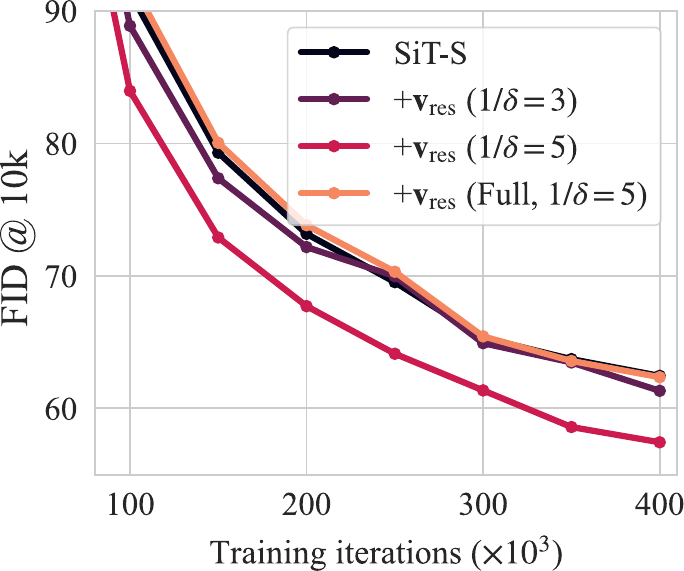}
        \caption{We compare the FID between models trained with the full and masked Jacobian regularization. The full residual regularization, with $1/\delta = 5$, behaves similarly to the masked residual using $1/\delta = 3$.}
        \label{fig:masked_vs_full_fid}
    \end{minipage}%
    \hfill
    \begin{minipage}[t]{0.54\linewidth}
        \centering
        \includegraphics[width=\linewidth]{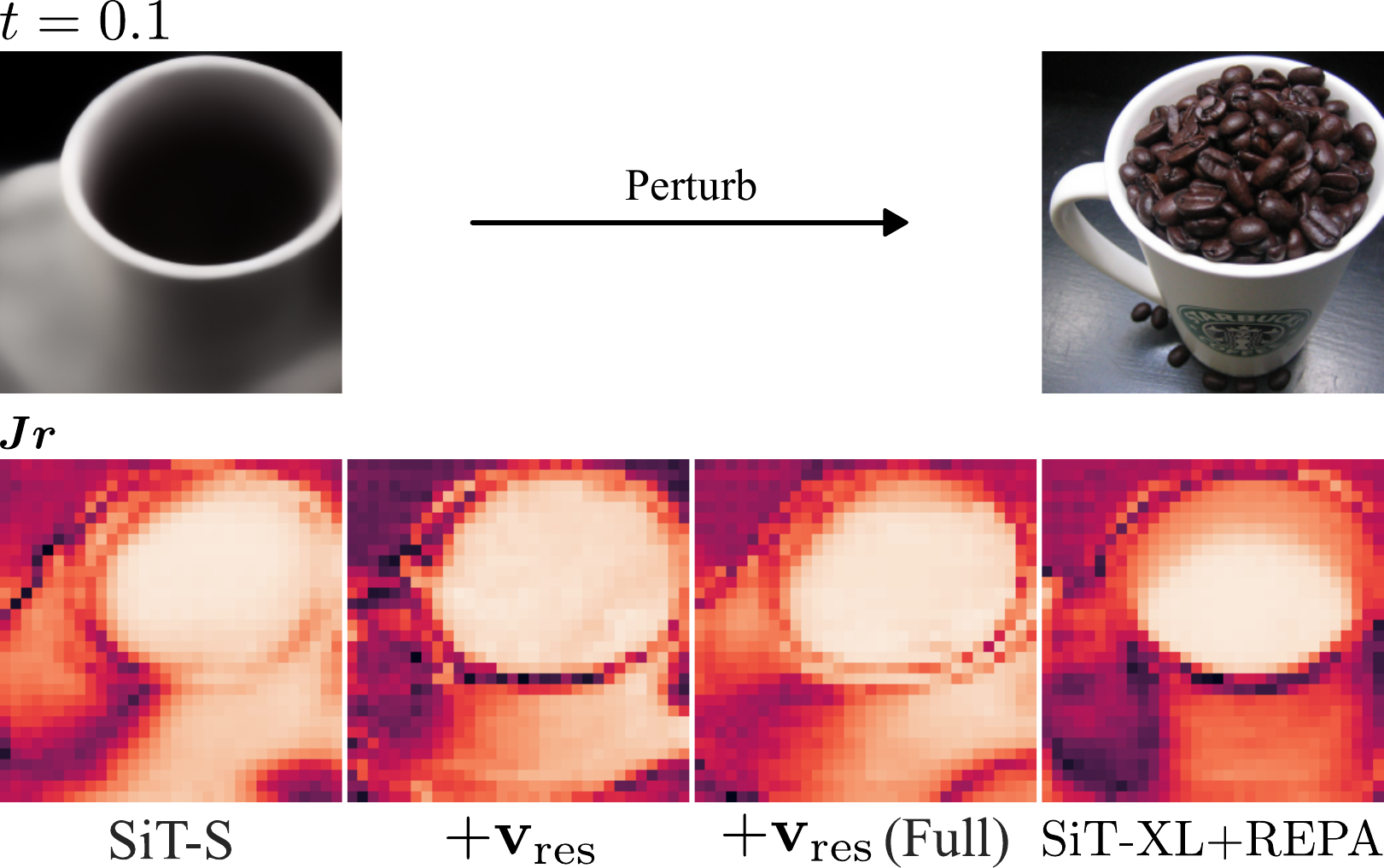}
        \caption{Qualitative comparison of the Jacobian-residual product. SiT-XL + REPA not only has larger eigenvalues (Figures \ref{fig:measurements} vs. \ref{fig:analysis_sit_s_full}) but also captures different spatial structures. While the base SiT-S and the SiT-S+Full are similar, the model trained with masking highlights noticeably different structures.}
        \label{fig:masked_vs_full_jacobian}
    \end{minipage}%
\end{figure}

When we compare the FID achieved by each of the models (Figure~\ref{fig:masked_vs_full_fid}), we see that the SiT-S regularized with the full residual performs similarly to the masked residual model using $1/\delta=3$. However, it would be naive to claim that the eigenvalue statistics alone determine generative performance; a model with a massive eigenvalue that has memorized images for each ImageNet class would not attain a competitive FID score. Thus, we hypothesize that qualitative differences also emerge in the larger and Jacobian-regularized models. 

In Figure~\ref{fig:masked_vs_full_jacobian}, we probe this hypothesis, showing different Jacobian-residual products. The best-performing model (SiT-XL + REPA) encodes plausible structures for the clean image in the direction of the residual, while the SiT-S and SiT-S + Full residual models produce blurrier results. The masked residual model differs from the other SiT-S variants, as it exhibits a result that is qualitatively closer to the SiT-XL structured Jacobian-residual product. This backs our overall suggestion that the spectral properties we discuss in this paper are but one of the directions from which we can approach the first-order properties of the denoiser.

\section{Direct Jacobian regularization}
\label{sec:direct_regularization}

\begin{figure}[t]
    \centering
    \includegraphics[width=1.\linewidth]{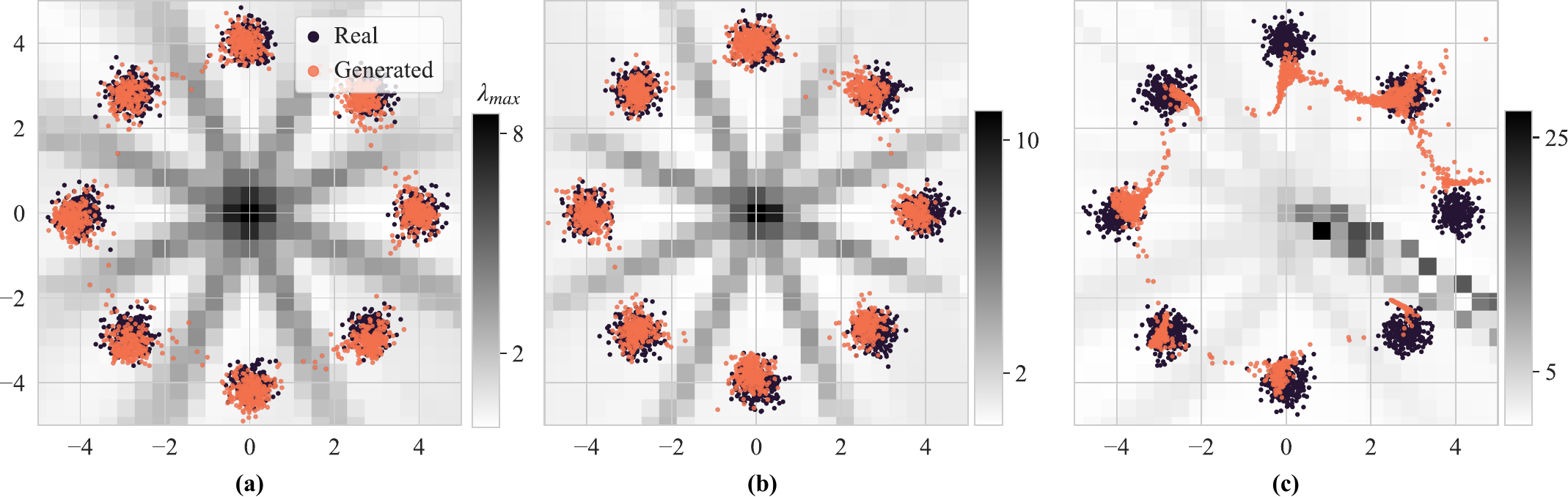}
    \caption{Training a denoising generative model on a mixture of 2D Gaussians. The grayscale color represents the maximum eigenvalue of the Jacobian at $t=0.5$ for each point on a 2D grid. 
    \textbf{(a)} Baseline model trained without regularization. \textbf{(b)} Jacobian regularization using the residual perturbation, pushing $\lambda_{\max}$ higher. \textbf{(c)} Direct regularization (Appendix~\ref{sec:direct_regularization}) maximizing all eigenvalues simultaneously. The resulting model is overly sensitive and fails to learn the target distribution.}
    \label{fig:toy2}
\end{figure}

\subsection{Training}
Instead of using the perturbed regression objective of Eq~(\ref{eq:residual_objective}) to regularize the Jacobian of the trained denoiser, we can directly use the finite-difference approximation of Eq~(\ref{eq:finite_difference}). By choosing the appropriate $\vv$ and the sign (minimize/maximize), we control the regularization, e.g., $\vv ~ \normal(\vzero,\mI)$ for suppressing diagonal directions or $\vv = -\vr$ to boost the eigenvalues. For the residual direction, we get the objective
\begin{equation}
    \mathcal{L}(\theta) = \lVert f_{\theta}(\vx_t,t) - \vx_0 \rVert_2^2
    + \tau \lVert f_{\theta}(\vx_t - \delta\vr,t) - f_{\theta}(\vx_t,t)  \rVert_2^2.
    \label{eq:direct_jac}
\end{equation}
We train an SiT-S model with this objective, using $1/\delta = 5$ and $\tau=0.0005$; for any larger $\tau$ we tried, the loss diverged. The resulting denoiser achieves an FID of $\sim$65, considerably higher than the 59 achieved by the baseline SiT. We hypothesize that this direct Jacobian regularization is more prone to noise in training and largely unstable. The regularization proposed in the main text is grounded in the base denoising regression objective, which, even when it doesn't provide any additional signal (e.g., when using a small $1/\delta$), does not hurt the overall training.

\subsection{Residual vs. Normal expansion}
\label{sec:contrast_random}
We can contrast the result of the residual direction, which we showed overlaps with eigenvectors proportionally to the corresponding eigenvalue (Eq.~\ref{eq:residual_overlap}), by analyzing the case where we maximize Jacobian expansion along random directions $\ve \sim \normal(\vzero, \mI)$. Using the same approach, we can analyze the covariance of the perturbation direction
\begin{equation}
    \mathbb{E}[ \ve \ve^T \mid \vx_t ] 
    = \mathbb{E}[ (\ve - \mathbf{0})(\ve - \mathbf{0})^T ]
    = \operatorname{Cov}(\ve) = \mI.
\end{equation}
If we project a random vector $\ve$ on the eigenvectors $\vv_i$ of $\mJ$ we would get
\begin{equation}
    \ve = \sum_i \vv_i^T \ve \vv_i = \sum_i q_i \vv_i
\end{equation}
which allows us to write the covariance as
\begin{equation}
    \mathbb{E}[ \ve\ve^T \mid \vx_t ] =
    \sum_i \sum_j \mathbb{E}[ q_i q_j \mid \vx_t ]\vv_i \vv_j^T.
\end{equation}
In this case, however, we have no identity that connects the covariance to the Jacobian, giving instead
\begin{equation}
    \sum_{i,j} \mathbb{E}[ q_i q_j \mid \vx_t ] \vv_i \vv_j^T = \mI = \sum_i 1\cdot \vv_i \vv_i^T
\end{equation}
which, by using the same RHS orthonormal basis argument, leads to
\begin{equation}
    \mathbb{E} [ q_i q_j \mid \vx_t ] = 1\ \Rightarrow\ \mathbb{E} [ q_i^2 \mid \vx_t ] = 1.
\end{equation}
Thus, using a randomly sampled Normal direction to maximize Jacobian expansion increases all eigenvalues simultaneously. In contrast, the residual direction focuses on the largest eigenvalues, which we hypothesize helps avoid introducing uncontrolled noise to the trained model, as it does not increase sensitivity in random directions. Furthermore, the proposed perturbed-input regularization does not naturally lead to this isotropic objective. Implementing this regularizer would require the direct Jacobian probing method, which we established before as subpar. We provide an example in Figure~\ref{fig:toy2}, showing how this isotropic maximization leads to a bad generative model.

\section{Toy experiment}
\label{sec:toy_experiment}

In the toy experiment, we train a simple MLP denoiser with three hidden 128-dim layers and SiLU activations. We encode the timestep with 64-dim sinusoidal embeddings, and use a zero-initialized output linear layer to map to the 2D $\vx_0$ prediction. Each model is trained for 10,000 iterations, with a learning rate of $0.002$ and a batch size of 512, using the x-prediction and $w(t)=1/(1-t)^2$ (equivalent to the v-loss formulation of \citet{li2026back}).

For the ``square'' regularization, we randomly sample $\vv = \{\pm [0,1]^T, \pm [1,1]^T\}$ and set $\delta=1$ and $\tau=0.1$. For the residual regularization, we use $1/\delta=10$ and $\tau = 1$. To visualize the largest eigenvalue, we use backwards differentiation to construct the $2\times2$ Jacobian matrix for every point and use SVD to get the largest singular value. As mentioned in our initial discussion, in all our experiments we consider the Jacobian to be symmetric, and thus use the singular value and eigenvalue terms interchangeably.

\section{ImageNet Training}
\label{sec:training_dynamics}

In Algorithm~\ref{alg:training_iteration} we describe a training iteration with the residual regularization. The regularization loss can only be computed \emph{after} the initial denoising prediction since it relies on the model output to construct the residual perturbation direction. For our ImageNet experiments, we construct the mask $\mM$ by keeping the top-5\% of pixels in the residual. For stochastic regularization, we follow the same training scheme, exchanging the residual direction with a randomly sampled Normal vector. 

We train all models using the same setup, setting the learning rate to $0.0002$, clipping gradient norms to $1.0$, and using a model EMA with decay $0.9999$. For the Rayleigh quotient estimate, we set the decay $\alpha=0.99$, roughly updating our estimates every 100 steps per $t$. We pre-extract VAE latents for all images, using the SD-1.5 VAE as with the base SiT models.

\begin{algorithm}[t]
\caption{Training iteration with Jacobian regularization}
    \textbf{Input}: Denoiser $\vf_{\theta}(\vx_t,t)$, sample $\vx_0$, 
    spectrum estimate $R_t$, EMA decay $\alpha$, regularizer weight $\tau$.
    \begin{algorithmic}[1]
        \LineComment{Denoising loss}
        \State $\vx_t = t\vx_0 + (1-t)\vepsilon,\ \vepsilon \sim \normal(\vzero,\mI)$
        \State $w(t) = 1/(1-t)^2$
        \State $\mathcal{L}_{\text{denoise}} = w(t) \left \lVert \vf_{\theta}(\vx_t,t) - \vx_0 \right \rVert_2^2$
        \LineComment{Jacobian Regularization}
        \State $\vr = \operatorname{sg}[\vf_{\theta}(\vx_t,t)] - \vx_0$ \Comment{Do not backprop through prediction}
        \If{masking}
            \State $\mM = \vone [ \vr \geq \mathcal{Q}_{0.95}(\vr)]$
            \State $\vr = \mM\vr$
        \EndIf
        \State $\delta_t = 1/(kR_t)$
        \State $\mathcal{L}_{\text{reg}} = w(t)
        \left \lVert \vf_{\theta}(\vx_t - \delta_t\vr,t) - \vx_0 \right \rVert_2^2$
        \LineComment{Update spectrum estimate}
        \State $\mJ_{\theta}\vr = (\vf_{\theta}(\vx_t - \delta\vr,t) - \vf_{\theta}(\vx_t,t))/(-\delta)$
        \State $R_t = \alpha R_t + (1-\alpha) (\vr^T\mJ_{\theta}\vr)/\lVert \vr \rVert_2^2$ 
    \end{algorithmic}
    \textbf{Return} $\mathcal{L} = \mathcal{L}_{\text{denoise}} + \tau\mathcal{L}_{\text{reg}}$
\label{alg:training_iteration}
\end{algorithm}

\begin{figure}[t]
    \centering
    \includegraphics[width=1.0\linewidth]{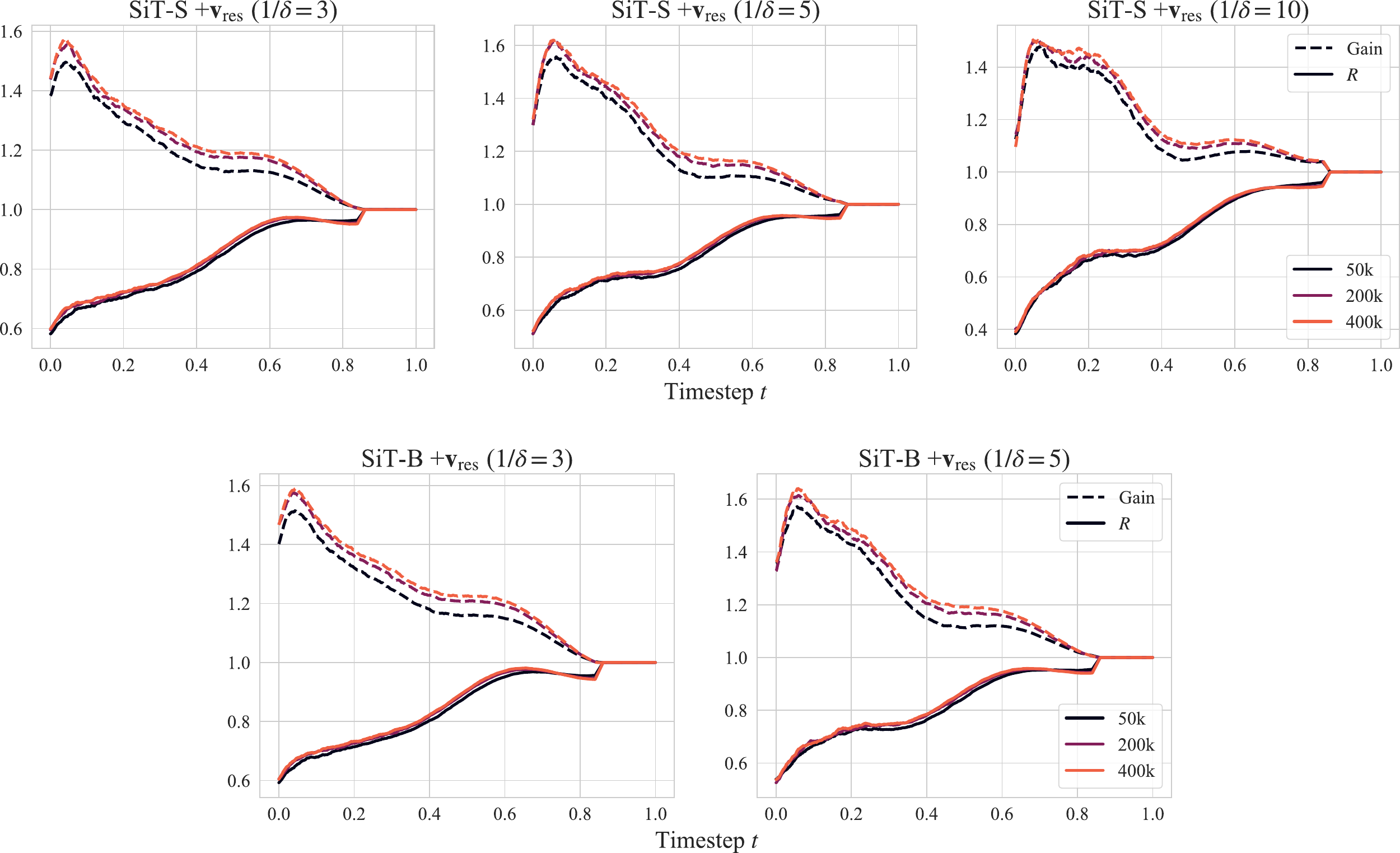}
    \caption{Gain and Rayleigh quotient $R$ during training for Jacobian-regularized SiT-S/B models. The gain overestimates the effect of the residual perturbation gain. We use $R$ to set the gain target during training.}
    \label{fig:gains}
\end{figure}

We choose the Rayleigh quotient for the eigenvalue target, instead of the raw gain $\lVert \mJ_{\theta}\vr \rVert/\lVert \vr \rVert$, as the Rayleigh quotient measures expansion parallel to the direction and ignores orthogonal changes. To visualize this difference, in Figure~\ref{fig:gains} we plot the gain $\lVert \mJ\vr \rVert/\lVert \vr \rVert$ and Rayleigh quotient $\vr^T\mJ\vr/\lVert \vr \rVert^2$ measured during the training of the Jacobian-regularized SiT-S/B. For numerical stability ($\lVert \vr \rVert \rightarrow 0$ as $t \rightarrow 1$), we set both to 1 for $t>0.85$. The gain overestimates expansion along the residual direction in early timesteps, where small changes in the direction of the residual have a large orthogonal component that is unrelated to the eigen-directions. Experimentally, we found the Rayleigh quotient to be the more stable training approach. 

Instead of modifying the target gain $1/\delta$, we initially increased $\tau$, and found the effect to be limited, as it only makes the regularization denoising have a larger gradient than the base denoising loss, which is dominated by the minimization of the residual. We set $\tau=1$ for all our residual Jacobian regularization experiments. For stochastic perturbations, $\tau$ is the main control signal, as the gain/Rayleigh quotient (which, in expectation, are equivalent) is fixed throughout training.

A dynamic that is important to mention is how the perturbed input $\vx_t - \delta\vr$ is easier for the denoiser to regress, since the residual $\vr = \vf_{\theta}(\vx_t,t) - \vx_0$ introduces the target image $\vx_0$ into the input. Using a larger $\delta$ makes the regularization task too simple, as we include more and more of the target $\vx_0$ in the input. On the other hand, by making $\delta$ too small, the task collapses to the base denoising loss. The online tuning we perform for the residual regularization is necessary to find the balance between simplifying the perturbed regression task and making it uninformative. In contrast to that, the stochastic regularization effect is stationary throughout training, and the response of the model to stochastic perturbations does not change as denoising improves.

\section{Visualizing eigenvectors}
\label{sec:viz_eigenvectors}

\begin{figure}[t]
    \centering
    \includegraphics[width=1.0\linewidth]{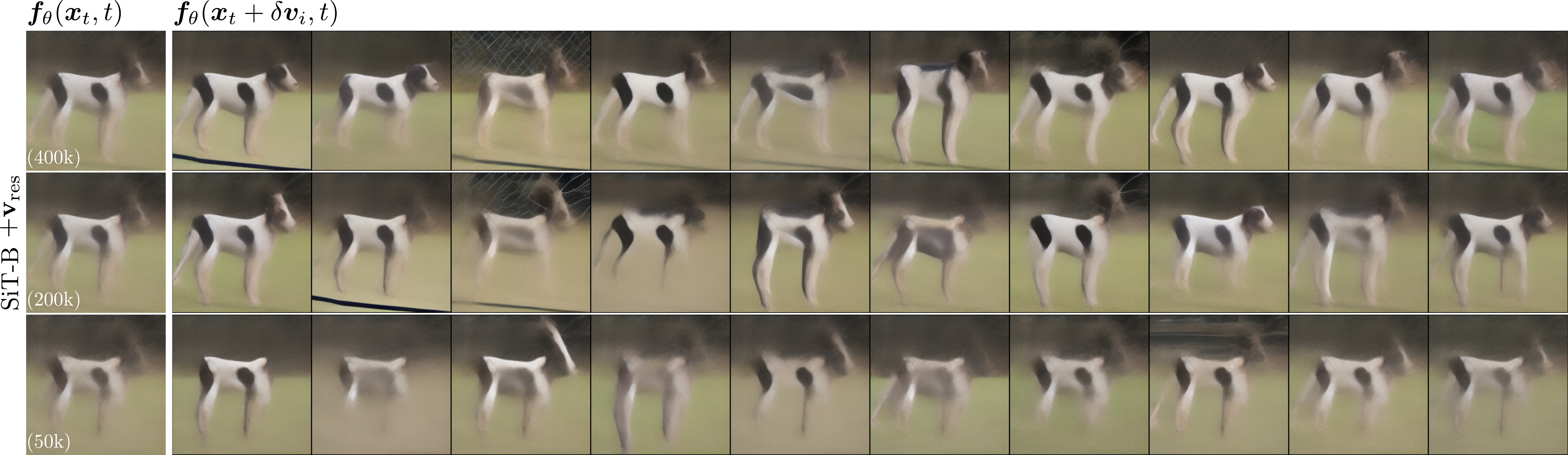}
    \textbf{(a)}
    \includegraphics[width=1.0\linewidth]{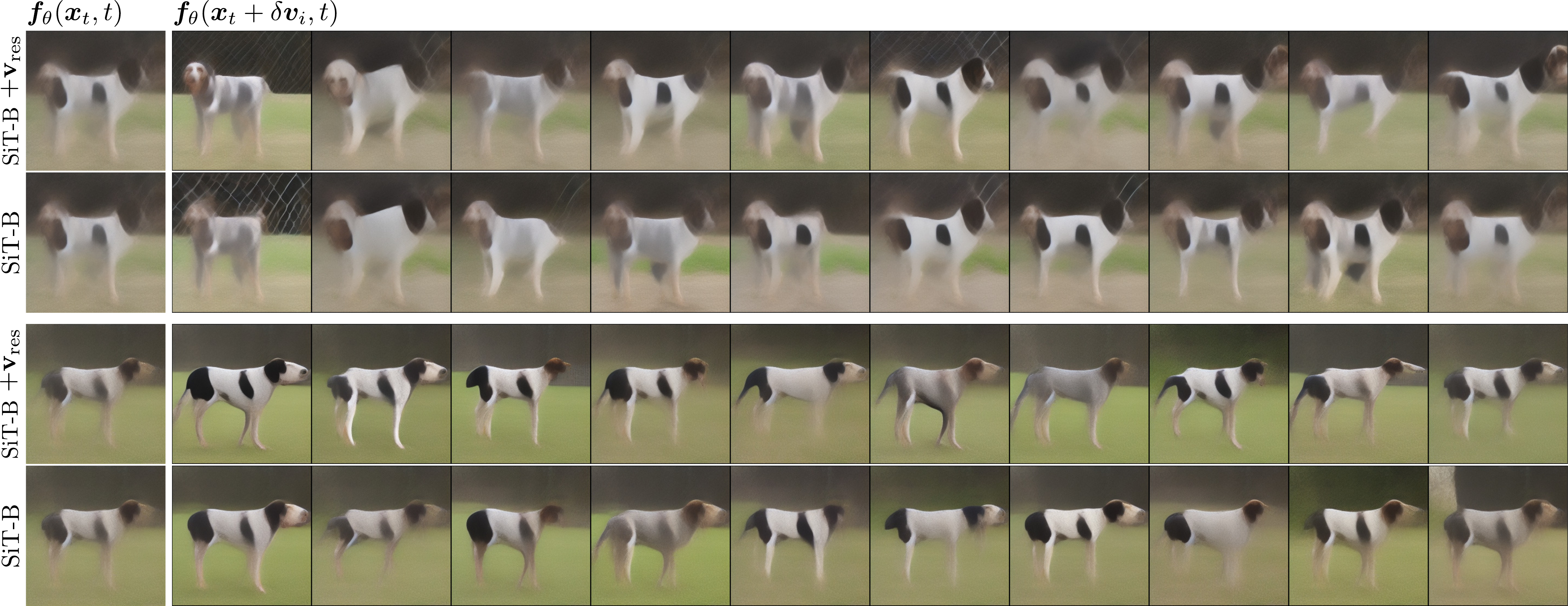}
    \textbf{(b)}
    \caption{Examples of a denoised image and the eigenvectors computed around it: \textbf{(a)} Different training iterations in the SiT-B + residual regularization model. As training progresses, the model captures wider and sharper variations in its top components.
    \textbf{(b)} The SiT-B model with and without residual regularization. The residual-regularized model improves the spectrum by making larger changes in the top components.}
    \label{fig:eigenvectors_trained}
\end{figure}

We visualize eigenvectors during training of the SiT-B model in Figure~\ref{fig:eigenvectors_trained} (a). During training, we can see the progression of the eigenvectors getting sharper and adding more variations to the directions. Moreover, we compare the eigenvectors for the baseline SiT-B and residual-regularized model in Figure~\ref{fig:eigenvectors_trained} (b), where although the difference is less prominent when comparing SiT-XL to SiT-S, we see similar patterns emerge.

The models we utilize all operate in the latent space of the SD-1.5 VAE \citep{rombach2022high}. To visualize the eigenvectors we compute, we must first decode them into image space, so that the changes that the eigenvector is pointing towards can be meaningfully shown as image pixels.

To decode latent eigenvectors to pixels, we first scale them by $s=100$, as decoding unit-norm latents through the decoder does not produce meaningful images, and also remove the decoder's zero-bias since decoding a zero latent does not correspond to a zero image. We can express that as $\mathcal{D}(\vf(100\vv_i, t)) - \mathcal{D}(\vzero)$, where $\mathcal{D}$ involves both the decoding and re-scaling to valid pixel values. In addition to eigenvectors, a more helpful visualization is to show the effect of perturbing the input towards the eigen-direction, i.e., $\vf(\vx_t + \delta\vv_i, t)$.

\begin{figure}[t]
    \centering
    \includegraphics[width=\linewidth]{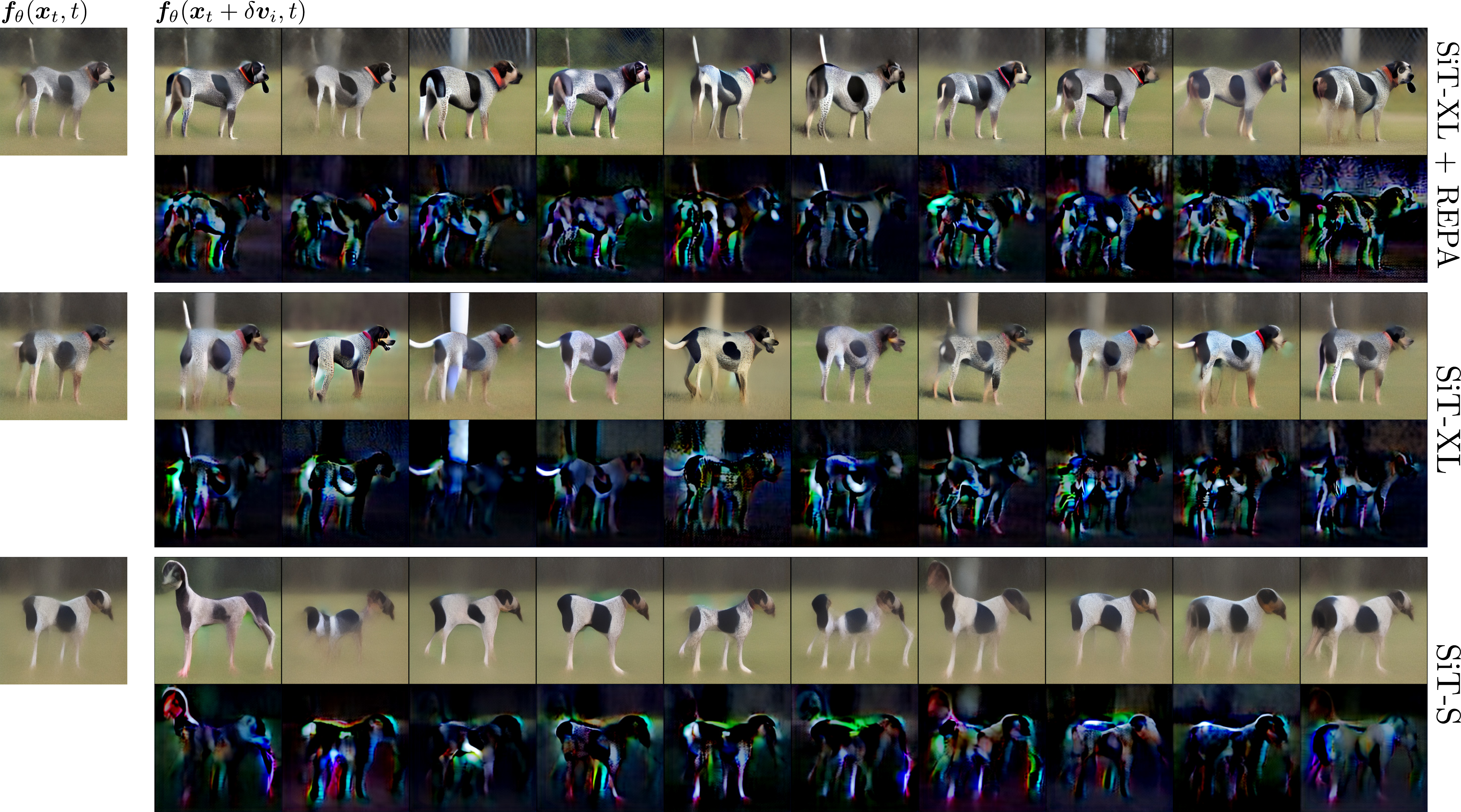}
    \textbf{(a)}
    \includegraphics[width=\linewidth]{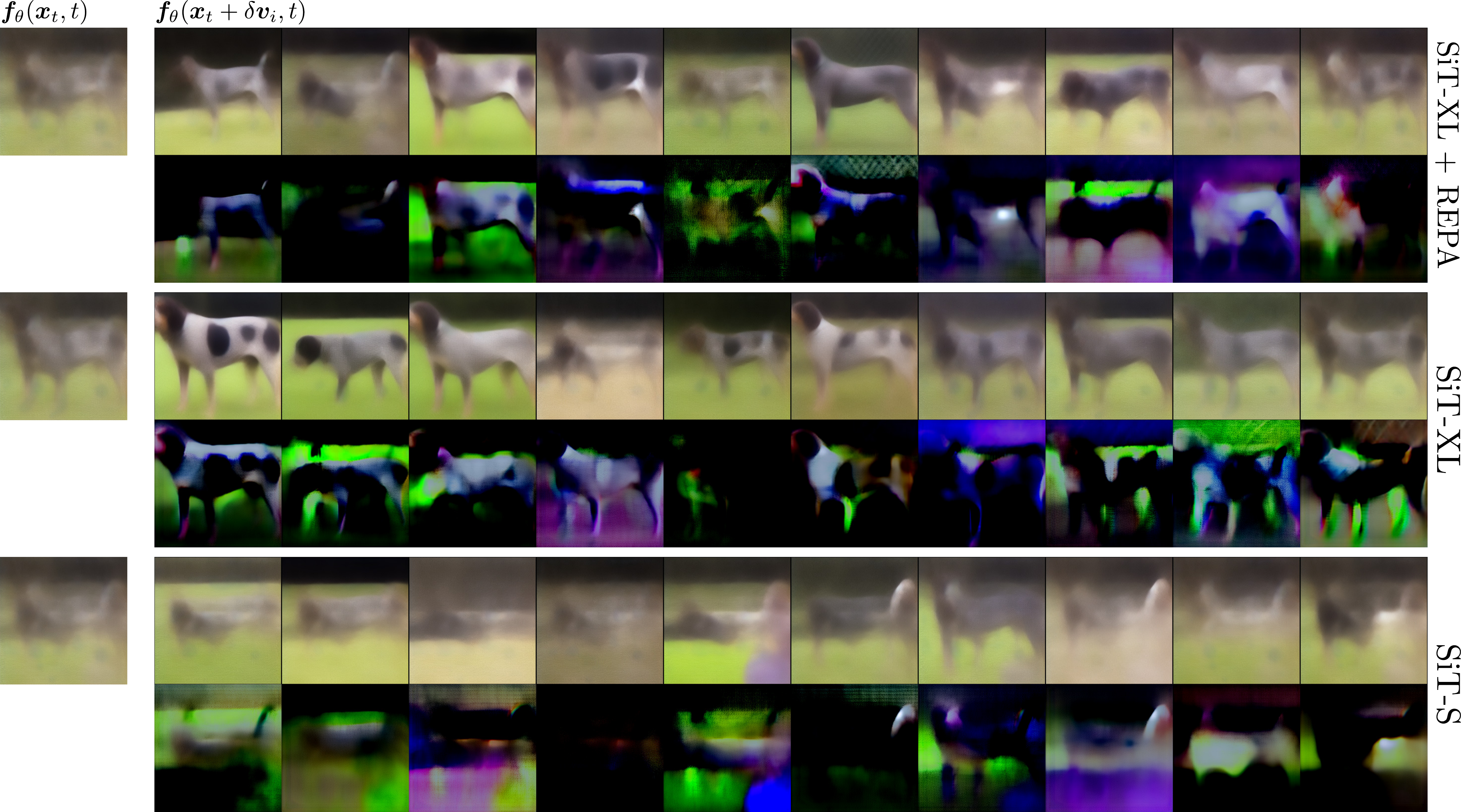}
    \textbf{(b)}
    \caption{\textbf{(a)} Eigenvectors at $t=0.4$, using classifier-free guidance with scale $w=4.0$. Increasing guidance leads to amplified differences between the eigenvectors and larger eigenvalues.
    \textbf{(b)} Eigenvectors at $t=0.2$ without classifier-free guidance. The variations in lower timesteps capture larger-scale structures in the images.}
    \label{fig:eigenvector_example_ots2}
\end{figure}

We provide examples of both the eigenvectors and the outputs of the denoiser in Figure~\ref{fig:eigenvector_example_ots2}. We first show eigenvectors at $t=0.4$ and guidance $w=4.0$, which produces larger eigenvectors for all models. We also visualize eigenvectors at $t=0.2$, where now the variations capture larger-scale structures in the image.

\section{Denoiser features}
\label{sec:feature_analysis}

\begin{figure}[t]
    \centering
    \includegraphics[width=\linewidth]{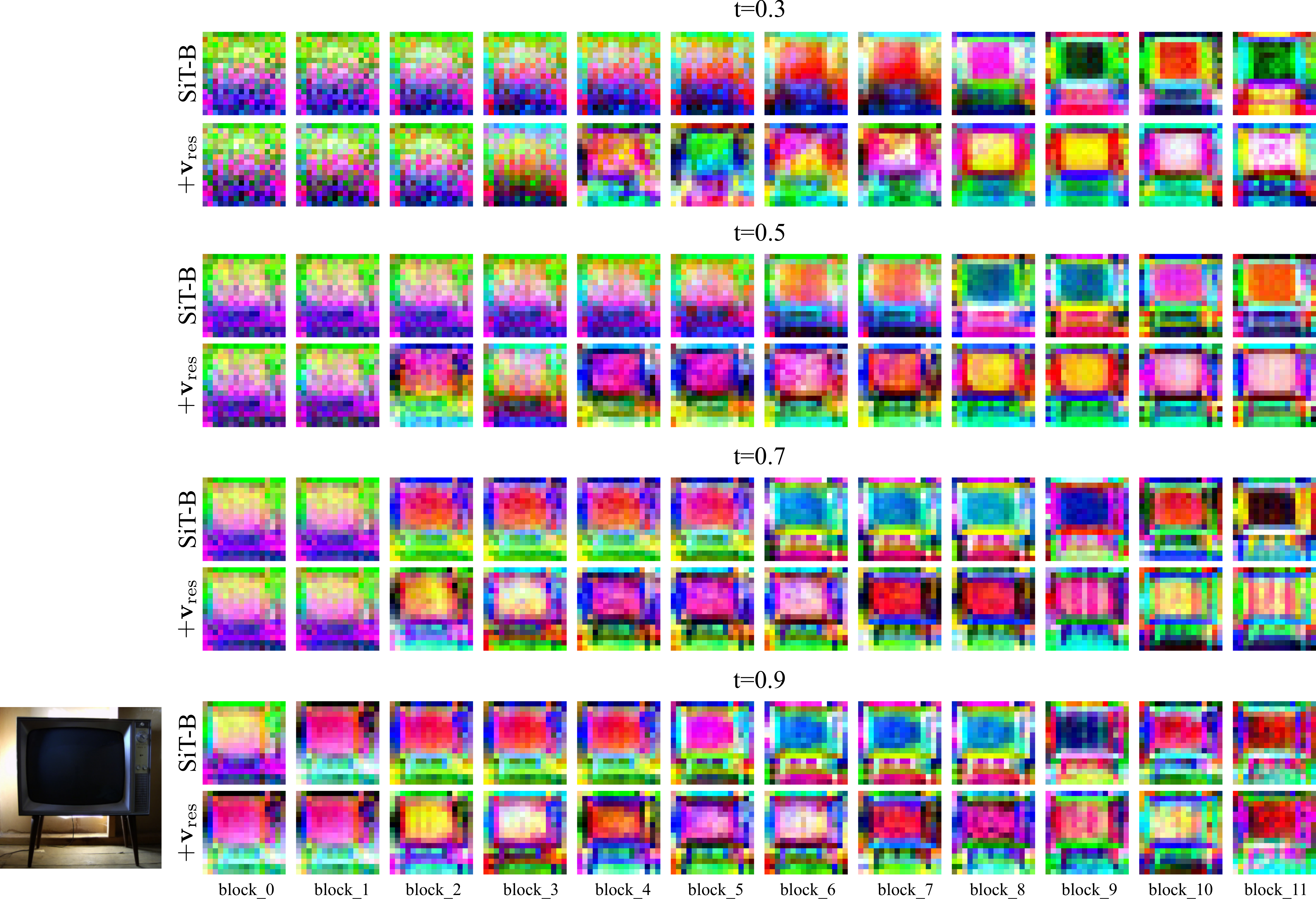}
    \\
    \vspace{1em}
    \includegraphics[width=\linewidth]{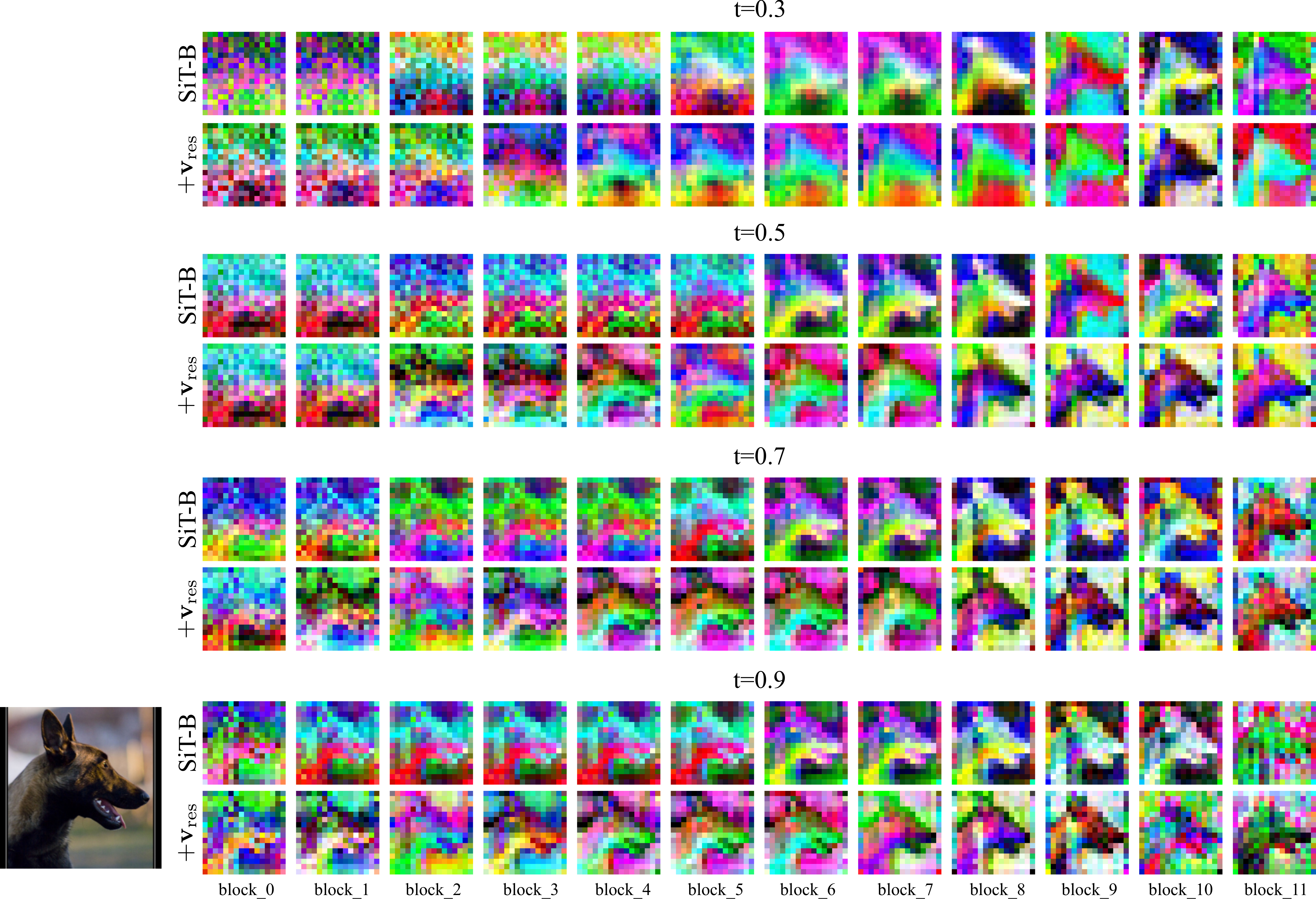}
    \caption{Feature comparison (reduced to RGB using PCA) between the baseline and the Jacobian-regularized SiT-B. At low timesteps ($t=\{0.3,0.5\}$), the image structures emerge in earlier blocks in the Jacobian-regularized model.}
    \label{fig:feats}
\end{figure}

In Figure~\ref{fig:feats}, we visualize the denoiser features for the base and the Jacobian-regularized SiT-B models. We $L_2$-normalize each transformer block's features and project them into RGB using PCA, fitting the principal components to each block separately. We observe that global structures seem to emerge in earlier blocks in the Jacobian-regularized model. For instance, in the $t=0.3$ row, the outline of the TV or the shape of the dog's head appears around block 3 for the Jacobian-regularized model, whereas it is not clearly visible until block 5 for the base SiT-B.

\end{document}